\documentclass[runningheads]{llncs}

\usepackage{eccv}

\usepackage{eccvabbrv}

\usepackage{graphicx}
\usepackage{booktabs}
\usepackage{caption}
\usepackage{multirow}
\usepackage[skins]{tcolorbox}
\usepackage[accsupp]{axessibility}  
\usepackage{tabularx}
\usepackage{soul}

\usepackage[pagebackref,breaklinks,colorlinks,citecolor=eccvblue]{hyperref}

\usepackage{orcidlink}

\newtcbox\fp{hbox, on line, colback=LimeGreen, enhanced, frame hidden, boxrule=0pt, top=-2pt, bottom=-2pt, right=-2pt, left=-2pt, sharp corners}
    
\newtcbox\secp{hbox, on line, colback=Goldenrod, enhanced, frame hidden, boxrule=0pt, top=-2pt, bottom=-2pt, right=-2pt, left=-2pt, sharp corners}

\newtcbox\tp{hbox, on line, colback=SkyBlue, enhanced, frame hidden, boxrule=0pt, top=-2pt, bottom=-2pt, right=-2pt, left=-2pt, sharp corners}
\newcommand{\myparagraph}[1]{\smallskip\noindent\textbf{#1.}}

\usepackage{subcaption} 

\makeatletter
\def\thanks#1{\protected@xdef\@thanks{\@thanks\protect\footnotetext{#1}}}
\makeatother

\begin{document}



\title{BAGS: Blur Agnostic Gaussian Splatting through Multi-Scale Kernel Modeling} 


\author{Cheng Peng$^{*}$\thanks{*Equal contribution.} \and
Yutao Tang$^{*}$ \and
Yifan Zhou \and
Nengyu Wang \and 
Xijun Liu \and
Deming Li \and
Rama Chellappa
}

\institute{Johns Hopkins University, Baltimore MD 21218, USA \\
\email{\{cpeng26,ytang67,yzhou223,nwang43,xliu253,dli90,rchella4\}@jhu.edu}}

\maketitle

\begin{abstract}
  Recent efforts in using 3D Gaussians for scene reconstruction and novel view synthesis can achieve impressive results on curated benchmarks; however, images captured in real life are often blurry. In this work, we analyze the robustness of Gaussian-Splatting-based methods against various image blur, such as motion blur, defocus blur, downscaling blur, \etc. Under these degradations, Gaussian-Splatting-based methods tend to overfit and produce worse results than Neural-Radiance-Field-based methods. To address this issue, we propose Blur Agnostic Gaussian Splatting (BAGS). BAGS introduces additional 2D modeling capacities such that a 3D-consistent and high quality scene can be reconstructed despite image-wise blur. Specifically, we
  model blur by estimating per-pixel convolution kernels from a Blur Proposal Network (BPN). BPN is designed to consider spatial, color, and depth variations of the scene to maximize modeling capacity. Additionally, BPN also proposes a quality-assessing mask, which indicates regions where blur occur. Finally, we introduce a coarse-to-fine kernel optimization scheme; this optimization scheme is fast and avoids sub-optimal solutions due to a sparse point cloud initialization, which often occurs when we apply Structure-from-Motion on blurry images.
  We demonstrate that BAGS achieves photorealistic renderings under various challenging blur conditions and imaging geometry, while significantly improving upon existing approaches. \footnote{Code: \href{https://github.com/snldmt/BAGS}{https://github.com/snldmt/BAGS}}
\end{abstract}

\section{Introduction}
\label{sec:intro}

High quality scene reconstruction and novel view synthesis from 2D images is a long-standing research problem with extensive applications in robotics, virtual reality, e-commerce, cinematography, \etc. Significant advancements in this field have been made in recent years with the introduction of Neural Radiance Field (NeRF)~\cite{DBLP:conf/eccv/MildenhallSTBRN20}. NeRF can achieve photorealistic view synthesis by implicitly representing the scene with a Multi-Layer Perceptron (MLP) and optimizing the MLP with differentiable ray-tracing. Many subsequent works have since been proposed to improve various aspects of NeRF, including optimization acceleration~\cite{SunSC22,fridovich2022plenoxels,mueller2022instant,tensorf}, anti-aliasing~\cite{barron2022mipnerf360, barron2023zip}, dynamic and generative modeling~\cite{pumarola2020d,tretschk2021nonrigid,liu2023robust,poole2022dreamfusion}, \etc. Recently, 3D Gaussian Splatting (3DGS)~\cite{3dgs} has been introduced as an alternative to NeRF-based approaches. 3DGS uses Gaussians as explicit 3D representations and a differentiable rasterization technique, which enables high quality scene reconstruction in a short time.


\begin{figure}[h]
\includegraphics[width=\linewidth]{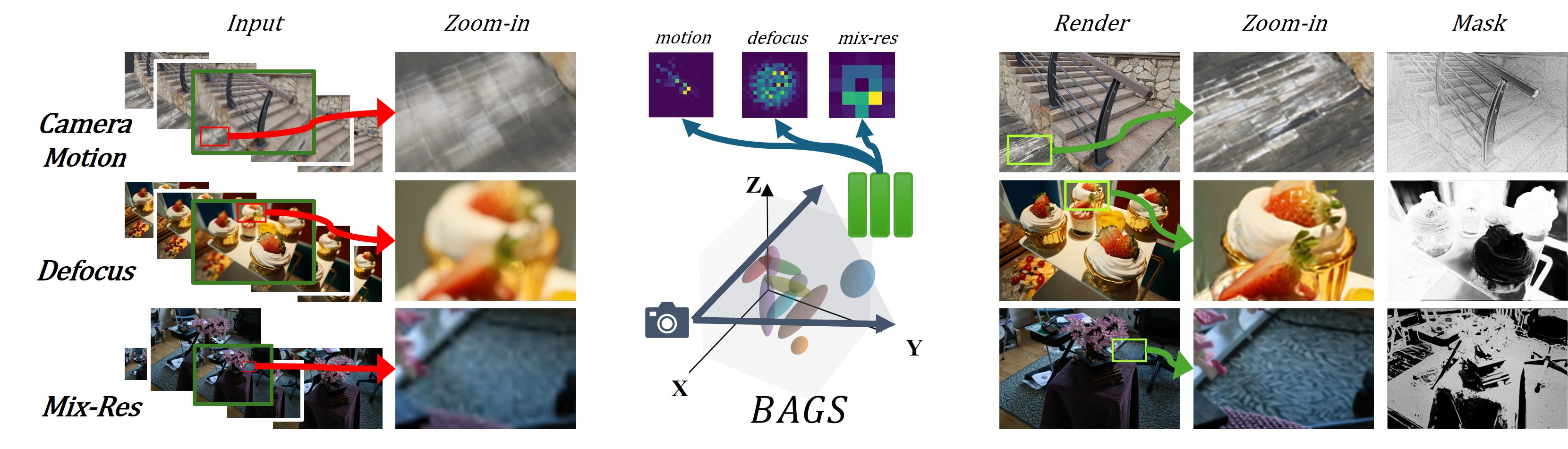}
\caption{We introduce BAGS, which can reconstruct high quality scenes even from blurry training images. Moreover, BAGS can provide kernels and masks that indicate the types and regions of the blur, as shown by the highlighted regions under Mask. }
\label{fig:intro}
\vspace{-2em}
\end{figure}

Practical applications of 3DGS, however, can be challenging. Despite its impressive performance on curated datasets, 3DGS requires high quality images and a good point cloud initialization to work well. Without these conditions, 3DGS often generates undesired Gaussians to overfit observation noise, leading to worse renderings, \eg, compared to NeRF. For instance, images acquired in real life or from the internet are often in non-ideal conditions. The camera may be unstable or out of focus, leading to motion or defocus blur. Images may also have different resolutions depending on the sensor types or post processes. 

There often exists sufficient information within a multi-view image set to generate a high quality reconstruction, even if the individual images are degraded. This is because the same region is often observed repeatedly from different views; such redundancy can then be explored to better recover the underlying scene. Works like Deblur-NeRF~\cite{ma2022deblur} and its variants~\cite{pdrf,dpnerf,badnerf,exblurf} are designed based on this insight. Formally, a simple yet powerful forward model can be used to describe the potentially blurry pixel value $\tilde{C}(x)$ at coordinate $x$:

\begin{equation}
\label{Eq:sparseblur}
    \tilde{C}(x) = \sum_{x_k\in \mathcal{N}(x)}C({x_k})h(x_k) \text{ s.t } \sum_{x_k\in \mathcal{N}(x)}h(x_k)=1,
\end{equation}
where $C(x_k)$ is the clear pixel value at pixel coordinates $x_k$ around the neighborhood of $x$, and $h(x_k)$ is the blur kernel weights. 
In this formulation, $C(x_k)$ is multi-view consistent, and $\tilde{C}(x)$ is multi-view inconsistent due to the blur from $h$. 
Previous methods~\cite{ma2022deblur,pdrf,dpnerf,exblurf,badnerf} aim to simultaneously optimize a multi-view consistent $C(x_k)$ through NeRF and the multi-view inconsistent $\tilde{C}(x)$ through a view-specific network that predicts $h(x_k)$. \textit{As such, we render only the clear $C(x_k)$ after training.} In comparison, vanilla NeRF \cite{DBLP:conf/eccv/MildenhallSTBRN20} treats $C(x_k)$ and $\tilde{C}(x)$ as one variable, forcing multi-view inconsistencies to be baked into the scene. 




In this work, we build upon the blur formulation in Eq. (\ref{Eq:sparseblur}), and explore two key differences between NeRF and 3DGS in robust reconstruction. NeRF-based approaches suffer from slow training speed, \ie only limited number of rays can be rendered at once. This is exacerbated by kernel modeling, which requires the rendering of a patch to model a single pixel. To mitigate this issue, current approaches model the kernel $h$ sparsely~\cite{ma2022deblur,pdrf,dpnerf} or only for specific types of blur~\cite{exblurf,badnerf}. Since 3DGS can rasterize entire images efficiently, it potentially has great synergy with Eq. (\ref{Eq:sparseblur}). On the flip side, NeRF requires only calibrated camera poses for reconstruction, whereas 3DGS also requires a good point cloud initialization. For noisy images, camera poses can often be estimated based on just a few correct matches, but a dense point cloud is challenging to obtain.

As shown in Fig~\ref{fig:intro}, we introduce Blur Agnostic Gaussian Splatting, or BAGS, which is a Splatting-based method that is robust against various types of blur. At the core of BAGS is a Blur Proposal Network (BPN). Contrary to previous methods~\cite{ma2022deblur,pdrf,dpnerf} that sparsely estimate $h$, BPN directly estimates dense kernels on full images during training. We find that estimating a dense $h$ is more effective and no less efficient than estimating a sparse one, which also has to predict kernel positions $x_k$. 
BPN considers the position and view embedding of pixel $x$, accounting for the spatial variance in image blur. Furthermore, a three-layer Convolution Neural Network (CNN) is used to compute the RGBD features of the rendered image. These extracted features allow BPN to consider the color and depth variation of the scene, \eg around edges and corners, more effectively.
Finally, BPN also computes a per-pixel mask, which blends the estimated clear and blurry pixels together. We make the observation that not necessarily all pixels are blurry; therefore, by constraining on its sparsity, such a mask can visualize regions where blur occurs. This is particularly useful as a way to evaluate image quality.

In addition to BPN, BAGS proposes a coarse-to-fine kernel optimization scheme. In some cases, BAGS cannot perform reconstruction well if 
it jointly optimizes BPN with a very sparse point cloud, due to the inherent ambiguity of separating blur from the observation. Furthermore, optimizing a dense kernel on full resolution directly is computationally expensive and time-consuming. 
To solve these conundrums, we first optimize the scene and BPN at lower scales with a smaller kernel dimension. This allows the optimization process to address a less challenging and ambiguous problem initially. We then gradually increase the resolution of the training images and the estimated kernels; since the blur problem has been partially addressed at a small scale, optimization at larger scales is more stable and effective. Under this scheme, we not only observe improved performance but also faster optimization process. 

In summary, our contributions can be summarized in three parts:

\begin{enumerate}
    \item We introduce a Blur Proposal Network, which considers spatial, depth, and color variations of the scene to model image blur; BPN can also produce a mask that indicates blurry regions in an input image.
    \item We introduce a coarse-to-fine optimization scheme, which gradually increases the training image resolution and the estimated kernel size with additional neural network layers; this improves the stability of the joint optimization process given a sparse point cloud.
    \item We evaluate the overall method, BAGS, on three image blur scenarios and find significant quantitative and visual improvements compared to current SoTA methods. 
\end{enumerate}


\section{Related Work and Background}
\myparagraph{Single Image Restoration} Restoring images from degradation, such as blur, has a long history of research. In general, this is a heavily ill-posed problem as many potential high quality images can lead to the same degraded observations. Traditionally, various priors are used to constrain the solution space~\cite{DBLP:journals/tip/ChanW98,RUDIN1992259,DBLP:conf/cvpr/KrishnanTF11,DBLP:conf/cvpr/XuZJ13,5674049,DBLP:conf/nips/Levin06}. More recently, deep-learning-based methods have achieved great performances by learning a better prior in a data driven way, e.g. as a function that directly maps blurry images to clean images. In this paradigm, we train a powerful restoration model with millions of training pairs and a large neural network. Ideally, images can be restored by such a model~\cite{cao2023physics,
yu2022towards,
lee2021iterative, cho2021rethinking, abuolaim2022improving, wang2022zero, kong2023efficient,Zamir2021MPRNet}. This paradigm has many issues in practice. For example, real training pairs are difficult to acquire in large scale; therefore, synthetic data is often used, which leads to significant domain shift problems. Every type of degradation also requires its own training sets and models, which exacerbate the data and computation costs. Even for degradations such as low resolution, for which is easy to synthesize training pairs, the restoration is unlikely to be multi-view consistent. 

\myparagraph{NeRF in Non-Ideal Conditions}
NeRF~\cite{DBLP:conf/eccv/MildenhallSTBRN20} works well when images are clean and well-calibrated; however, images acquired in the wild are often less ideal. Factors such as low light, camera motion, object motion, and incorrect focus can degrade image quality or affect multi-view consistency. Many works introduce domain knowledge to model the non-ideal contributors, such that the radiance field only models a canonical, multi-view consistent scene. RawNeRF~\cite{Mildenhall_2022_CVPR} uses an approximated tonemapped loss and variable exposure to account for low light observations. NeRF-W~\cite{DBLP:conf/cvpr/Martin-BruallaR21} uses an image-specific embedding to model inconsistent appearances and transient objects. For object motion, many works~\cite{park2021nerfies,park2021hypernerf,DBLP:conf/cvpr/PumarolaCPM21,tretschk2021nonrigid} use an implicit function to describe scene deformation, such that the traced rays are first deformed.
For image degradation such as blur, similar ideas have been used~\cite{ma2022deblur,dpnerf,badnerf,pdrf,hdrnerf} to model the forward image degradation process; \ie the 3D consistent scene and 2D inconsistent degradation will be separated based on the multi-view photometric loss. 


\myparagraph{3D Gaussian Splatting}
Recently, 3DGS~\cite{3dgs} introduces a novel, point-based scene reconstruction method that improves upon NeRF in many aspects. 3DGS utilizes a set of 3D Gaussian primitives~\cite{zwicker2001surface} $\{\mathcal{G}_n|n=1,...,N\}$ to represent the scene. These Gaussians are parameterized as follows:
\begin{equation}
\label{Eq:gaussian}
    \mathcal{G}_n(\textbf{v}) = e^{-\frac{1}{2}(\textbf{v}-\textbf{p}_n)^{T}\Sigma^{-1}_{n}(\textbf{v}-\textbf{p}_n)},
\end{equation}
where the center $\textbf{p}_n\in \mathbb{R}^{3 \times 1}$ and covariance matrix $\Sigma_n\in \mathbb{R}^{3 \times 3}$ define the position, scale, and rotation of the Gaussian at point $\textbf{v}\in \mathbb{R}^{3 \times 1}$ in space. Additionally, each Gaussian also contains opacity $\alpha$ and color $c$. For rendering, the Gaussians are projected onto the image plane as $\mathcal{G}^{\textrm{2D}}_n$, where the center and covariance matrices are transformed based on the camera rotation $\textbf{R}\in \mathbb{R}^{3 \times 3}$ and translation $\textbf{t}\in \mathbb{R}^{3 \times 1}$:
\begin{equation}
\label{Eq:gaussian2d}
    \textbf{p}^{\textrm{2D}}_n = \textbf{R}\textbf{p}_n + \textbf{t}, \Sigma^{\textrm{2D}}_n = \textbf{R}\Sigma_n\textbf{R}^{T}.
\end{equation}
We can then aggregate 2D Gaussians according to their depth order $D$ over a pixel/tile, based on alpha blending:

\begin{equation}
\label{gsrender}
\begin{gathered}
    C(x)=\sum_{i=1}^{D} T_i\alpha_i\textbf{c}_i\mathcal{G}^{\textrm{2D}}_i,
    T_i = \prod_{j=1}^{i-1}(1-\alpha_j\mathcal{G}^{\textrm{2D}}_j).
\end{gathered}
\end{equation}
With well-designed tiling, this splatting-based rasterization technique is significantly faster than ray-tracing in NeRF-based methods without affecting rendering quality; various work have ensued to further improve the Gaussian representation~\cite{yang2023deformable,compressedgs,navaneet2023compact3d,gao2023relightable} and the rasterization process~\cite{mipsplatting,scaffoldgs}.

\section{Blur Agnostic Gaussian Splatting} 
\begin{figure}[h]
\includegraphics[width=\linewidth]{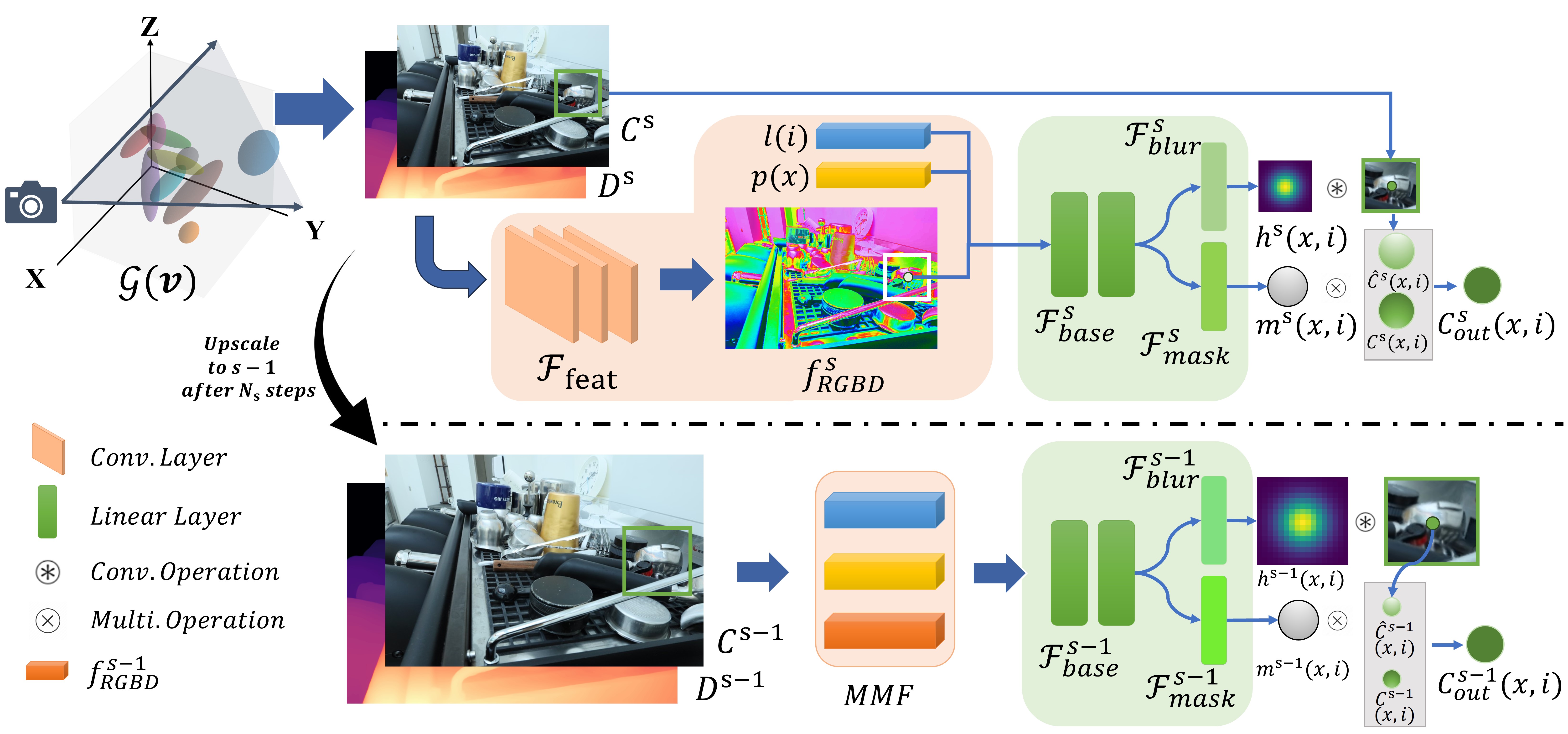}
\caption{BAGS is optimized by training a Blur Proposal Network on top of the scene $\mathcal{G}(\textbf{v})$ over multiple scales. \textbf{Top:} We extract $f^s_\textrm{RGBD}$ from color and depth, which are concatenated with the position and view embedding $p(x), l(i)$ to form the Multi-Modal Feature (MMF). The kernel MLP then estimates the per-pixel kernel $h^s$ and mask $m^s$. We use $h^s$ to model the blur image $\tilde{C}^s$ and employ $m^s$ to blend the rendered image $C^s$ and the blur-modeled image $\tilde{C}^s$, yielding ${C}_{\textrm{out}}^s$.
\textbf{Bottom:} After $N_s$ steps, we upscale image resolution and modify the kernel MLP to produce $h^{s-1}$ with a larger kernel size.}

\label{fig:method}
\vspace{-2em}
\end{figure}

BAGS is designed to robustly optimize 3D Gaussians by introducing additional modeling capacities in 2D, which allows BAGS to model away 3D-inconsistent blur from the scene. As shown in Fig.~\ref{fig:method}, BAGS consists of two parts: a Blur Proposal Network (BPN) and a coarse-to-fine optimization scheme. For clarity, we first describe the design of BPN on a single scale, and then describe modifications on BPN and the training loss on scale $s$ for multi-scale training.

\subsection{Blur Proposal Network}

Following the forward blur model described in Eq. (\ref{Eq:sparseblur}), BPN models a per-pixel convolution kernel $h(x) \in \mathbb{R}^{K\times K}$ for every pixel $x$, where $K$ is the kernel size. 
The rasterization efficiency in Gaussian Splatting allows us to model $h(x)$ as a full convolution kernel. 
Additionally, BPN estimates a per-pixel scalar mask $m(x) \in [0,1]$, controlling the areas where blur modeling takes place.


\myparagraph{Multi-Modal Features} In scene reconstruction, a large neural network will impact training time. As such, we use a four-layer MLP $\mathcal{F}_{\textrm{kernel}}$ to model blur. To maximize the capacity of $\mathcal{F}_{\textrm{kernel}}$, we design Multi-Modal Features (MMF) that can help distinguish the scene from blur. Real image blur typically varies between different observed images and different pixels within an image; therefore, we provide $\mathcal{F}_{\textrm{kernel}}$ with a learnable view embedding $l(i)$ and a positional embedding $p(x)$ for training view $i$ and coordinate $x$. Additionally, we provide $\mathcal{F}_{\textrm{kernel}}$ with a set of features $f_\textrm{RGBD}$ extracted from the rendered image $C$ and depth $D$ from rasterization, using a small, three-layer CNN $\mathcal{F}_\textrm{feat}$. These features are concatenated with $l(i)$ and $p(x)$ to form the MMF for $\mathcal{F}_{\textrm{kernel}}$. Consequently, the forward process to estimate the kernel $h(x,i)$ and the mask $m(x,i)$ given a specific pixel $x$ and training view $i$ can be noted as:

\begin{equation}
\label{Eq:BPN}
\begin{gathered}
    h(x,i), m(x,i) = \mathcal{F}_\textrm{kernel}(l(i)\oplus p(x)\oplus f_\textrm{RGBD}(x,i)), \\
    \textrm{where } f_\textrm{RGBD}(x,i) = \mathcal{F}_\textrm{feat}(C(x,i) \oplus D(x,i)), 0 \leq m \leq 1.
\end{gathered}
\end{equation}

Incorporating RGBD features enbales BPN to explore the patch statistics around the current pixel, especially at edges and corners where blur is prominent. Scene depth also strongly correlates with the magnitude of blur. For example, defocus blur affects pixels that are outside of a specific focus plane; scene content that is further away from this focus plane becomes more blurry. For camera motion blur, pixels in the near plane may shift more than pixels in the far plane. We find the inclusion of $f_\textrm{RGBD}$ notably improves BPN's modeling capacity and the quality of novel view synthesis.

\myparagraph{Residual Mask Bottleneck} As described in Eq. (\ref{Eq:BPN}), $\mathcal{F}_\textrm{kernel}$ also estimates a per-pixel scalar mask $m(x,i)$. Specifically, this mask is used to blend the Gaussian-splatted image $C$ with the blur-modeled image $\tilde{C}$ as follows:

\begin{equation}
\label{eq:mask}
    {C}_{\textrm{out}}(x,i) = \left(1-m(x,i) \right)C(x,i) + m(x,i) \tilde{C}(x,i),
\end{equation}
where ${C}_{\textrm{out}}$ is the final output at training. We note that not necessarily all pixels are blurry in an observed image. The blending $m$ provides an additional degree of freedom to allow $C$, the clear pixel, to be directly compared with ${C}_{\textrm{obs}}$;
\ie, if ${C}_{\textrm{obs}}{(x,i)}$ suffers from blur, then $m(x,i)$ should be large, giving a higher weight to $\tilde{C}(x,i)$; if ${C}_{\textrm{obs}}(x,i)$ if free of blur, $m(x,i)$ should be small, favoring $C(x,i)$.


This design is similar to residual learning by providing the output with a skip connection from $C$ as initialization. We apply a small sparsity constraint on $m$ to encourage a larger weight towards $C$ by default. After training, we can extract $m$ to automatically identify the degraded regions in an input image. 




\vspace{-0.5em}
\begin{figure}[!htb]
    \setlength{\abovecaptionskip}{3pt}
    \setlength{\tabcolsep}{1pt}
    \begin{tabular}[b]{cccc}
        \begin{subfigure}[b]{.245\linewidth}
            \includegraphics[width=\textwidth]{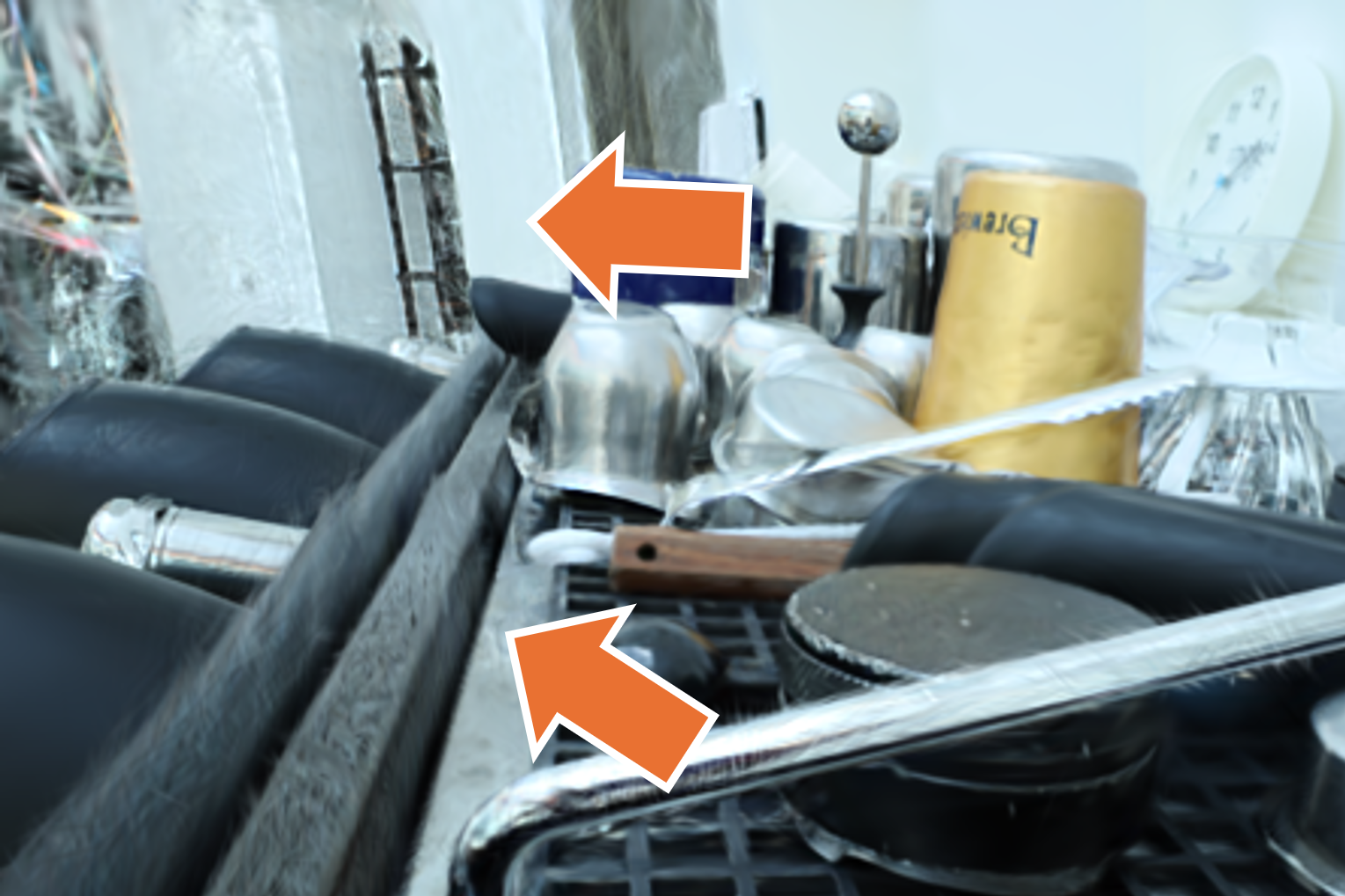}
            \caption{Without C2F}
            \label{wo_c2f}
        \end{subfigure} &
        \begin{subfigure}[b]{.245\linewidth}
            \includegraphics[width=\textwidth]{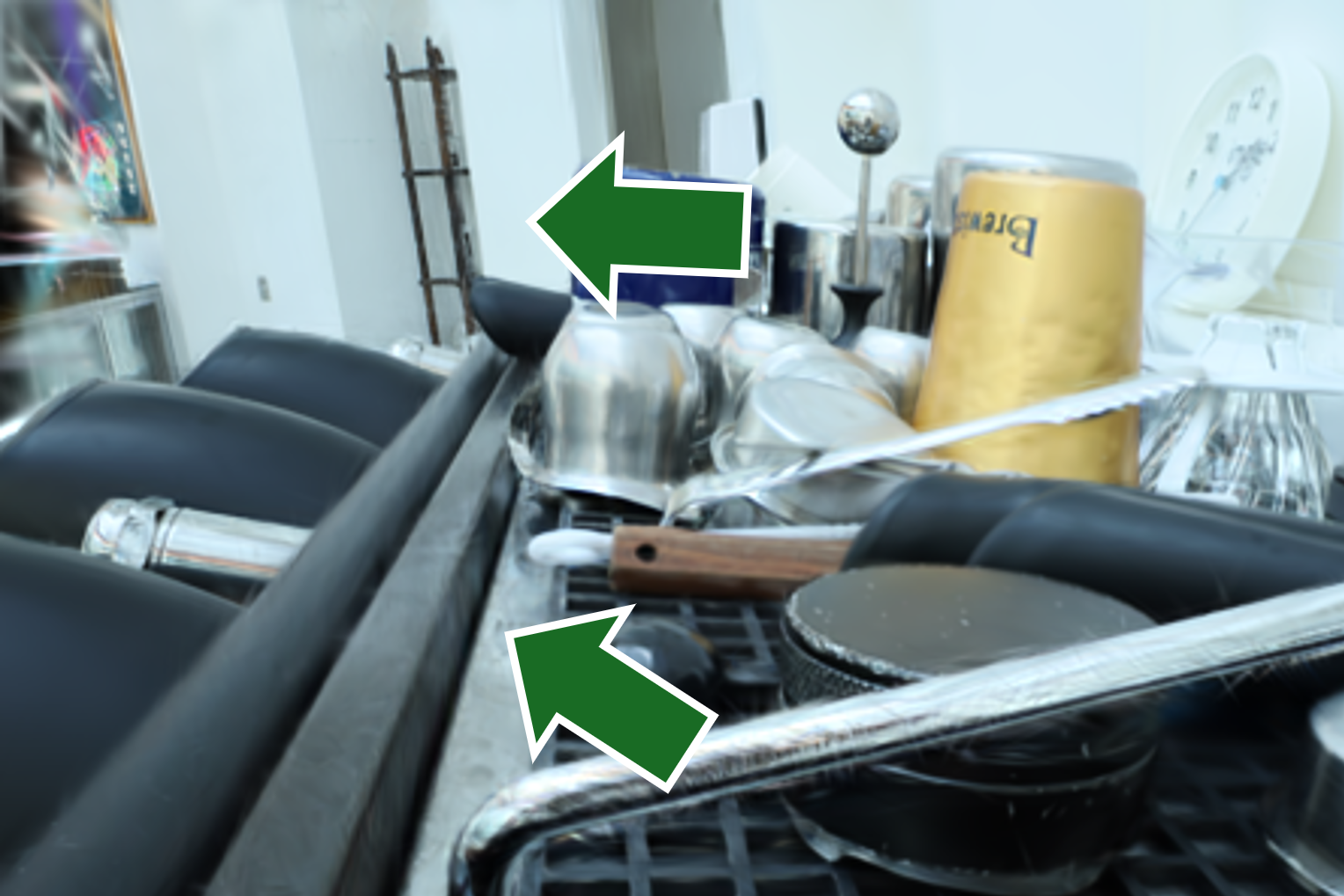}
            \caption{BAGS}
            \label{rob-gs-a}
        \end{subfigure} &
        \begin{subfigure}[b]{.245\linewidth}
            \includegraphics[width=\textwidth]{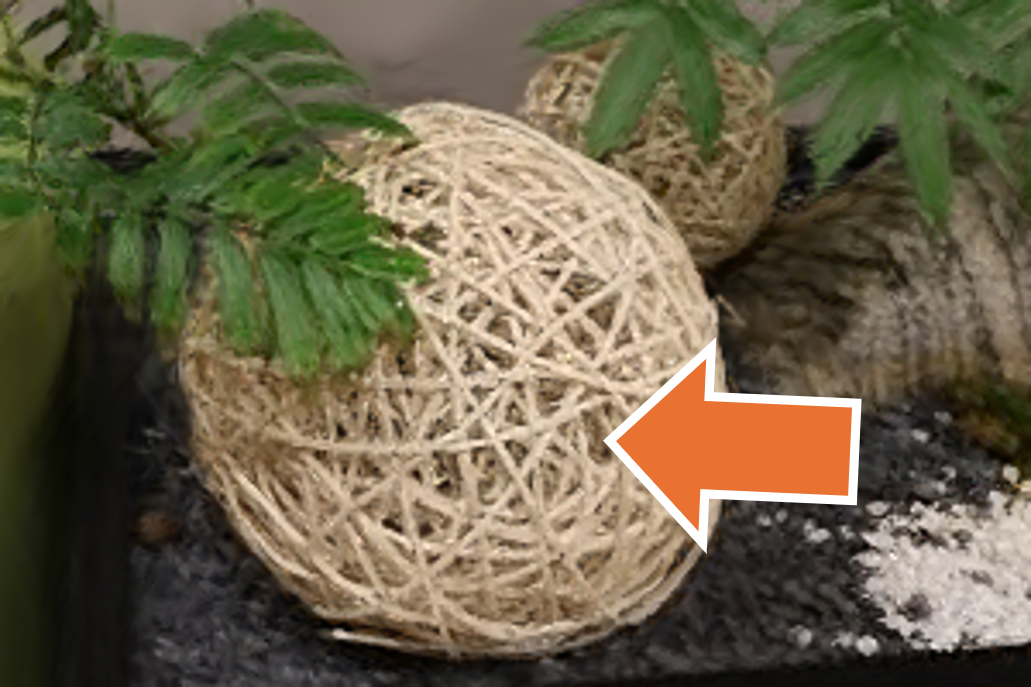}
            \caption{Naive densification}
            \label{naive}
        \end{subfigure} &
        \begin{subfigure}[b]{.245\linewidth}
            \includegraphics[width=\textwidth]{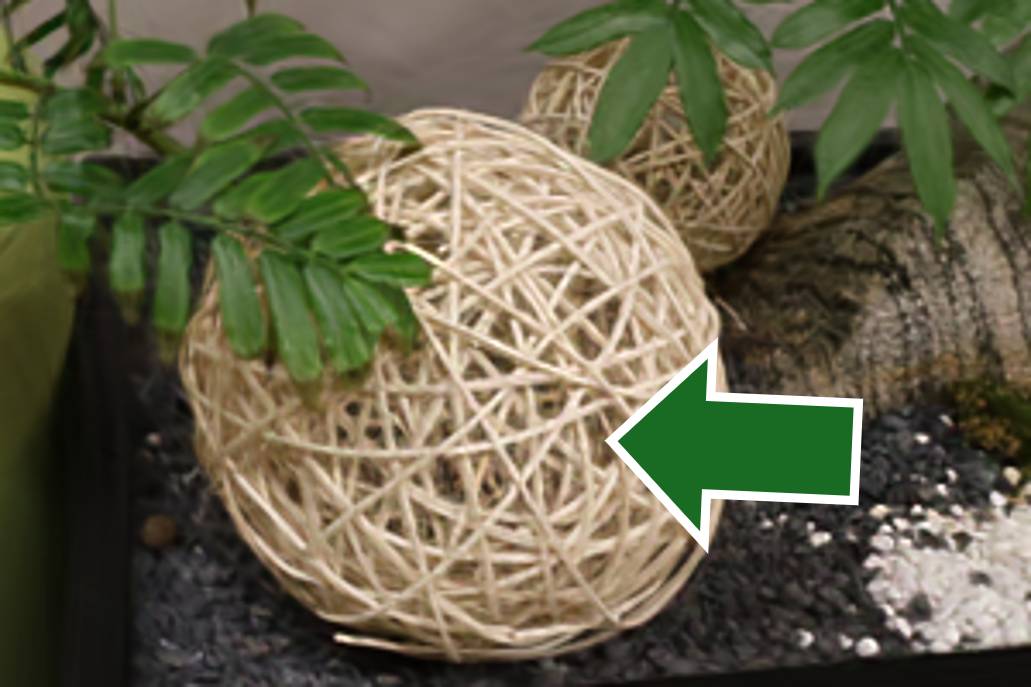}
            \caption{BAGS}
            \label{rob-gs-b}
        \end{subfigure}
    \end{tabular}
    \caption{Gaussians may get stuck at a local minimum without proper initialization. As indicated in \ref{wo_c2f}, optimizing $h$ directly can lead to noisy surfaces. Naively densifying the scene before adding $h$ can also lead to noisy Gaussians; as shown in \ref{naive}, the noisy Gasussians are not well removed even after adding $h$. By using a coarse-to-fine training schedule, we achieve better results in \ref{rob-gs-a} and \ref{rob-gs-b}.}
    \label{fig:method_vis}
    \vspace{-1em}
\end{figure}

\subsection{Coarse-to-Fine Kernel Optimization}

A high quality 3DGS reconstruction requires a good point cloud estimation from Structure from Motion (SfM). Camera calibration parameters, which can be estimated from just a few matches, are relatively robust against image blur. On the other hand, obtaining a sufficiently dense point cloud can be challenging. Unfortunately, 3DGS is prone to over-fitting without a good initialization.
This problem can worsen when a blur model is introduced and the kernels $h$ provide additional degrees of freedom in scene optimization, as shown in Fig.~\ref{fig:method_vis}. Furthermore, estimating dense per-pixel kernels on a full resolution image can be computationally expensive. A sparse estimation of the kernel does not meaningfully improve efficiency, as rendering is cheap in rasterization while the neural network to predict any type of kernel is still costly. Fundamentally, the complexity to compute per-pixel kernel scales quadratically with resolution. 
To address these problems, we propose a coarse-to-fine kernel optimization scheme, which initially optimizes Eq. (\ref{Eq:BPN}) at a low resolution and gradually increases render resolution, achieving a more stable training process and better efficiency.

\myparagraph{Multi-Scale BPN} For multi-scale optimization, BPN should generate kernels with a consistent field of view. We define scale based on the downscaling ratio $2^{s-1}$ in resolution, where the original resolution is defined as $s=1$. Additionally, if the image resolution is downscaled by a factor of two, \ie $s=2$, the proposed kernel size $K^s$ should also be similarly downscaled. As shown in Fig.~\ref{fig:method}, a proper $K^s$ ensures that the kernel is always modeling blur from the same regions. 

To enable multi-scale training, we separate $\mathcal{F}^s_{\textrm{kernel}}$ into two parts: a two-layer base MLP $\mathcal{F}^s_{\textrm{base}}$ and two one-layer MLP heads $\mathcal{F}^s_{\textrm{blur}}$ and $\mathcal{F}^s_{\textrm{mask}}$. The base MLP extracts features from MMF, such features are then provided to the two heads for generating the kernels $h^s$ and masks $m^s$ at scale $s$. In practice, only $\mathcal{F}^s_{\textrm{blur}}$ has a scale-dependent parameter size corresponding to the kernel size $K^s$; therefore, a new MLP layer with output dimension $K^s\times K^s$ is added for every scale. 
The updated multi-scale BPN can be described as follows:

\begin{equation}
\label{Eq:msBPN}
\begin{gathered}
    f_{\textrm{inter}}(x,i) = \mathcal{F}^s_{\textrm{base}}(l(i)\oplus p(x)\oplus f_\textrm{RGBD}(x,i))\\
    h^s(x,i) = \textrm{softmax}(\mathcal{F}^s_{\textrm{blur}}(f_{\textrm{inter}}(x,i))),\\
    m^s(x,i) = \textrm{sigmoid}(\mathcal{F}^s_{\textrm{mask}}(f_{\textrm{inter}}(x,i))).
\end{gathered}
\end{equation}


\myparagraph{Coarse-to-Fine Optimization}
We split training into multiple stages based on scale. Starting from the coarsest scale $s=S$ to the finest scale $s=1$, we optimize BAGS with a downsampling factor of $2^{s-1}$ applied to $C, \tilde{C}, {C}_\textrm{obs}$, and $h$. The optimization for each scale lasts for $N_s$ steps, we then increase the resolution in $C, \tilde{C}, {C}_\textrm{obs}$, and $h$. The scale-dependent training loss can be described as:

\begin{equation}
\label{render}
\mathcal{L}_s = \lambda_{\textrm{photo}}\lVert {C}^s_{\textrm{out}}-{C}^s_{\textrm{obs}}\rVert + \lambda_{\textrm{DS}}\mathcal{L}_{\textrm{D-SSIM}}({C}^s_{\textrm{out}}, {C}^s_{\textrm{obs}}) + \lambda_{\textrm{mask}}\lVert m^s\rVert,
\end{equation}
where ${C}^s_{\textrm{obs}}$ and ${C}^s_{\textrm{out}}$ are the observed and predicted image at scale $s$, ${\mathcal{L}}_{\textrm{D-SSIM}}$ is the structural similarity loss; $\lVert m^s\rVert$ is the mask sparsity loss. 
We find this training scheme achieves the best of both worlds: activating BPN at low resolution generates less overfitting artifacts during densifying the point cloud given the reduced visiblity of blur. Moreover, it also serves as an efficient warm-up for BPN at higher resolution, as computing per-pixel kernels at low resolution is much cheaper.


\section{Experiments}
We evaluate BAGS under three degradation scenarios: camera motion blur, defocus blur, and mix resolution. All of these degradations are commonly observed in images acquired in real life or online. We further introduce a new unbounded 360 drone dataset, which is collected at sunset or night and experiences degradation due to a combination of motion blur and low light condition.


\myparagraph{Camera Motion and Defocus Blur} Camera motion and defocus blur are often observed in real acquisitions, \eg due to low light, incorrect auto-focus, or limitation in the finite aperture of real cameras. Deblur-NeRF \cite{ma2022deblur} provides real scene acquisitions that contain camera motion and defocus blur; both blur categories consist of ten scenes. While real world scenes have blur-free references, the camera settings, \eg, exposure, may not be consistent with source views. Experiments on synthetic blur are reported in supplemental material.

\myparagraph{Mix Resolution} To analyze the scenario where the input images are of inconsistent resolution, we apply various downsampling operations on the Mip-NeRF 360 dataset~\cite{barron2022mipnerf360}.
Specifically, we split the training images into four equal parts and apply 4X, 3X, 2X, and no downscaling to each of the parts. The image splits are sampled in uniform randomness to ensure no downscaling covers a consecutive interval of the camera trajectory. 
We then use NeRFStudio's data processing pipeline~\cite{nerfstudio}, which converts all images to the highest resolution and calibrates with COLMAP~\cite{schoenberger2016sfm}. Similar to previous works~\cite{barron2022mipnerf360, barron2023zip, mipsplatting}, we perform the entire process on downscaled data instead of the original 4K resolution. 

\begin{figure}[!htb]
    \centering
    \setlength{\abovecaptionskip}{3pt}
    \setlength{\tabcolsep}{2pt}
    \begin{tabular}[b]{ccc}
        \begin{subfigure}[b]{.48\linewidth}
            \includegraphics[width=\textwidth]{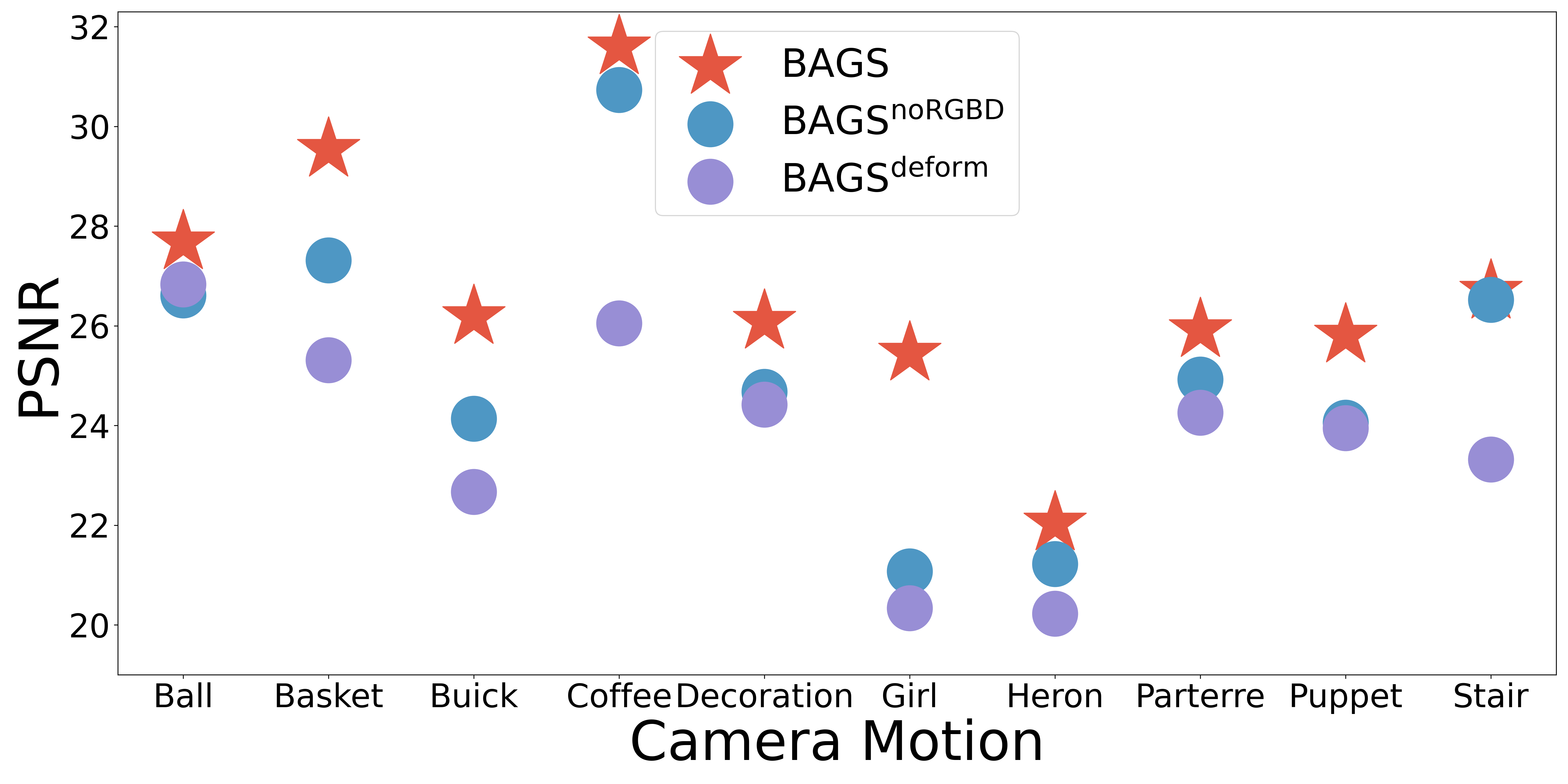}
            \caption{}
            \label{img:rgbd_deform}
        \end{subfigure} &

        \begin{subfigure}[b]{.48\linewidth}
            \includegraphics[width=\textwidth]{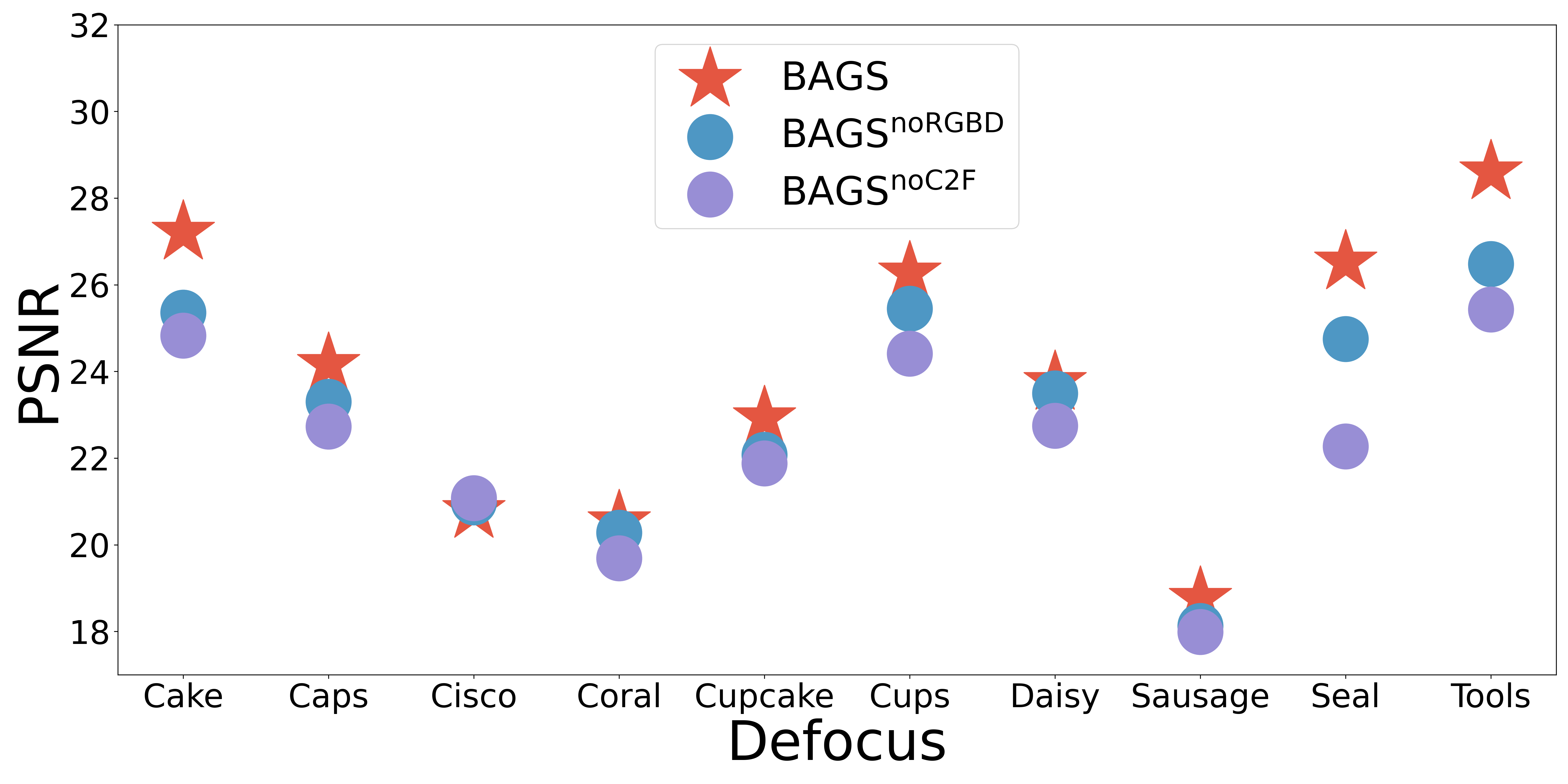}
            \caption{}
            \label{img:defocus}
        \end{subfigure} \\
        
        \begin{subfigure}[b]{.48\linewidth}
            \includegraphics[width=\textwidth]{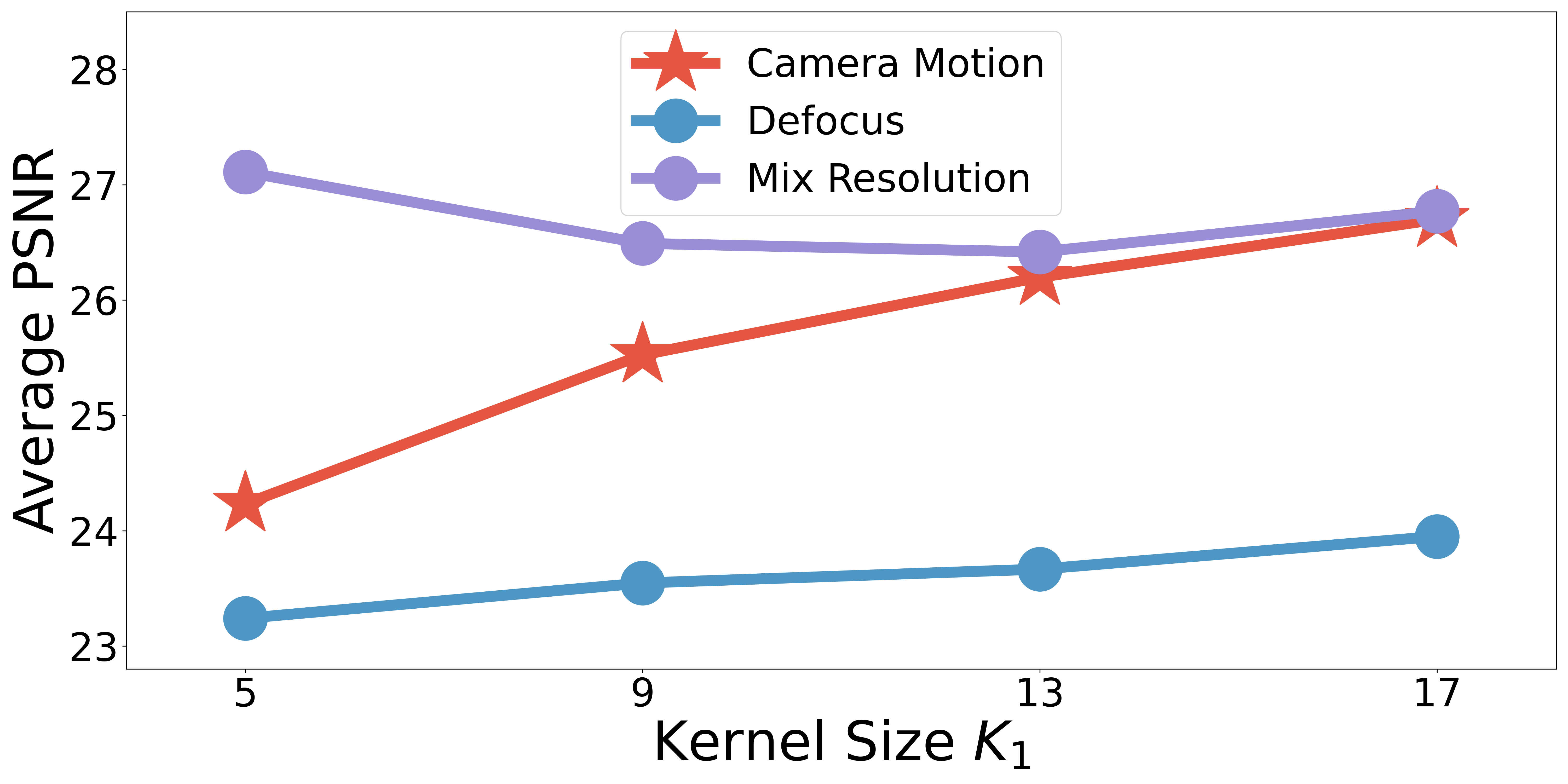}
            \caption{}
            \label{img:kernel_size}
        \end{subfigure} &
        \begin{subfigure}[b]{.48\linewidth}
            \includegraphics[width=\textwidth]{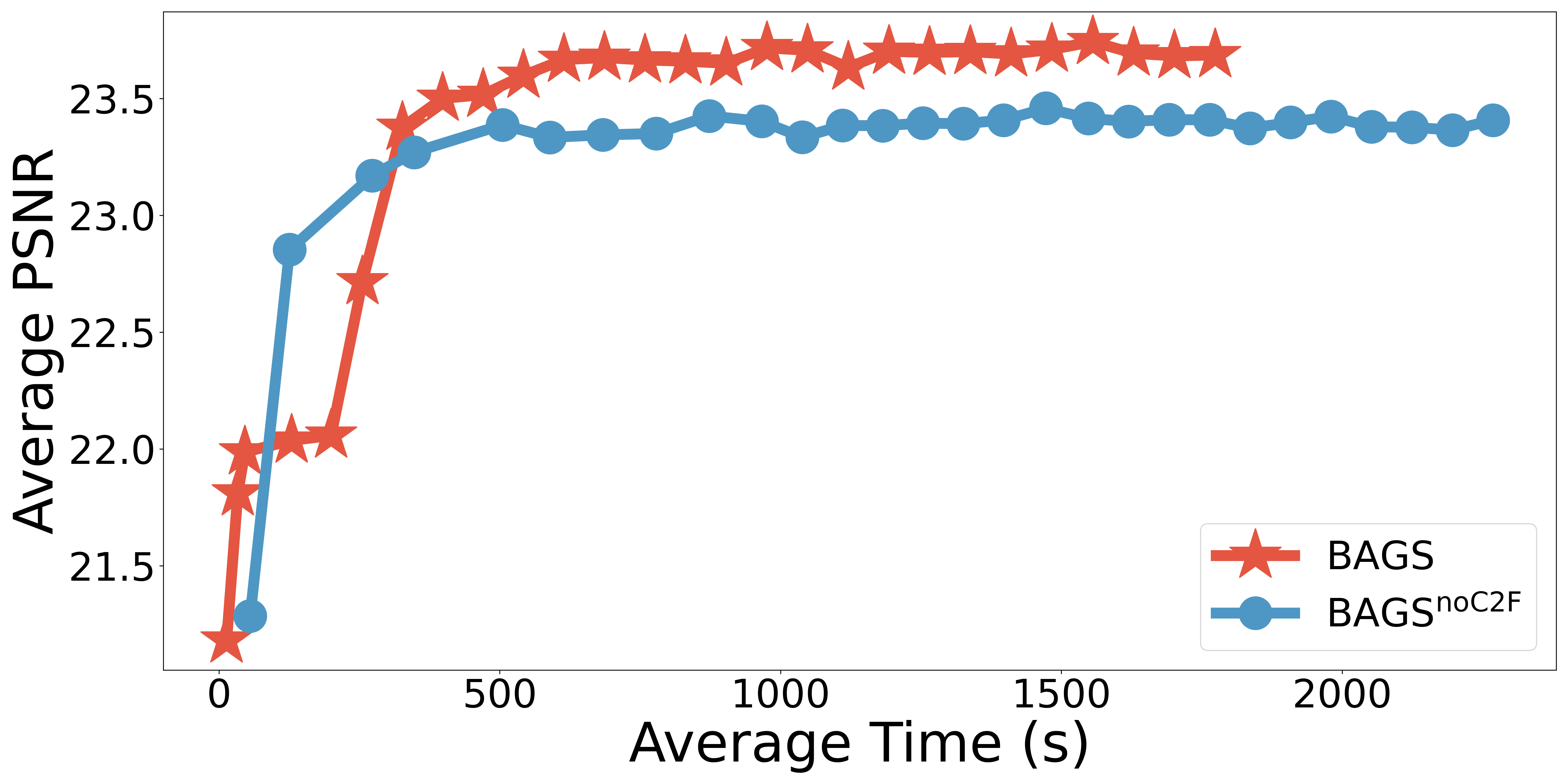}
            \caption{}
            \label{img:speed}
        \end{subfigure}

    \end{tabular}
    \caption{Ablation study of BAGS with different sub-modules. In~\ref{img:rgbd_deform} and~\ref{img:defocus}, introducing RGBD features and coarse-to-fine optimization improve novel view synthesis quality, and alternative sparse deformable kernel suffers in performance;~\ref{img:speed} demonstrates our speed improvement against $\textrm{BAGS}^{\textrm{noC2F}}$. In~\ref{img:kernel_size}, larger kernel generally leads to better performance, other than in the mix resolution scenario. }
    \label{fig:ablation}

\vspace{-1.5em}
\end{figure}

\myparagraph{Unbounded Drone Motion Blur} 
The dataset introduced by DeblurNeRF~\cite{ma2022deblur} serves as a good benchmark for performance comparison; however, it is collected in a forward bounded format. These forward bounded scenes focus on a small area and have significant overlaps within the images; as such, it is unclear if the methodology~\cite{ma2022deblur,pdrf,dpnerf,badnerf,exblurf} based on these sets can generalize to other scenarios. In fact, none of the deblurring methods can work for unbounded scenes. To supplement DeblurNeRF's dataset and understand how BAGS performs in realistic settings, we collect five sets of drone footage on different buildings in 360 unbounded format. These footages are collected at sunset or night, and suffer from camera motion blur and low light noise. For each scene, we also capture clear frames from a stable drone position for testing, resulting in around 100 images overall. We use NeRFStudio~\cite{nerfstudio} to calibrate images from these footages. More details on this dataset can be found in supplemental material.

\myparagraph{Implementation Details}
BPN's $\mathcal{F}_\textrm{feat}$ is constructed as a three-layer CNN; the intermediate channel size is 64, the kernel size for each layer is 5, 5, and 3. $\mathcal{F}_\textrm{kernel}$ is constructed with four linear layers; the intermediate channel size is 64. For multi-scale training, we use three scales starting with scale $s=3$, where $K_3=5$, $K_2=9$, and $K_1=17$. BAGS implements its Gaussian Splatting and rasterization based on Mip-Splatting~\cite{mipsplatting}, which has been shown to yield good results across multiple resolutions. More implementation details can be found in supplemental material.

\subsection{Ablation Studies}
We compare three designs in BAGS to demonstrate the effectiveness of different components: 

\begin{itemize}
    \item $\textrm{BAGS}^{\textrm{noRGBD}}$: We implement BPN without the RGBD features $f_{\textrm{RGBD}}$.
    \item $\textrm{BAGS}^{\textrm{deform}}$: We implement BPN with a sparse deformable kernel, similar to DeblurNeRF~\cite{ma2022deblur}.
    \item $\textrm{BAGS}^{\textrm{noC2F}}$: We implement BAGS without a coarse-to-fine kernel optimization scheme.
\end{itemize}
We further investigate the effect of the final kernel size $K_1$ on reconstruction quality across all degradation scenarios. 

As shown in Fig.~\ref{fig:ablation}, the introduction of RGBD features to BPN greatly improves its ability to disentangle image blur from the underlying scene. The addition of $f_{\textrm{RGBD}}$ increases the average performance by 1.57/0.92 dB in PSNR and 0.03/0.03 in SSIM for camera motion and defocus blur, respectively. This agrees with our intuition that blur is highly correlated with depth and image features like edges. Our coarse-to-fine optimization scheme also notably boosts robustness against defocus blur by 1.64dB and 0.08 in PSNR and SSIM on average. As visualized in Fig.~\ref{wo_c2f}, directly optimizing 3DGS on a sparse point cloud in $\textrm{BAGS}^{\textrm{noC2F}}$ leaves many ambiguities; while training views have low loss, test views show a lack of the correct surface structure and many artifact Gaussians. Densifying Gaussians without BPN first improves performance for defocus blur; however, this strategy is bad for motion blur, as shown in Fig.~\ref{naive}. In Fig.~\ref{img:speed}, we show that our coarse-to-fine strategy is not only performant, but also fast; on average, BAGS can reach high performance in less than fifteen minutes, which is much faster than the two hours required by PDRF~\cite{pdrf}. This strategy is also faster compared to single-scale training, as full-scale training in our scheme uses less iterations, and lower scale training does not sacrifice performance. Both multi-scale and single-scale training take 40K iterations overall.

We also implement a deformable version of BPN in $\textrm{BAGS}^{\textrm{deform}}$, where BPN estimates the kernel position and kernel weights for ten points within a $20\times 20$-pixel patch. We show its performances in Fig.~\ref{img:rgbd_deform}, where $\textrm{BAGS}^{\textrm{deform}}$ is tested on camera motion blur. Camera motion blur often has a sparse representation, \ie the camera motion trajectory~\cite{badnerf,exblurf}; therefore, a ten-point estimation of the kernel may perform well. However, we find that $\textrm{BAGS}^{\textrm{deform}}$ still underperforms compared to a full kernel estimation in BAGS. This may be due to the difference between NeRF and 3DGS; \ie, a ray-based deformation can be better expressed in NeRF, while a patch-based convolution can be better expressed in 3DGS. We also note that $\textrm{BAGS}^{\textrm{deform}}$ is not more efficient than BAGS despite its sparse estimation. This is because approaches like~\cite{ma2022deblur,dpnerf,pdrf} require one forward pass to estimate one kernel position and weight; therefore, we need to forward multiple times for every pixel to estimate the sparse kernel. In comparison, BAGS only requires a single forward pass to obtain the full kernel. 

\begin{table*}[htbp]
    \centering
    \resizebox{\linewidth}{!}{

    \begin{tabular}{l |cccccccccccccccccc}

     Camera  & \multicolumn{3}{c}{NeRF \cite{DBLP:conf/eccv/MildenhallSTBRN20}}  & \multicolumn{3}{c}{Mip-Splatting \cite{mipsplatting}} &  \multicolumn{3}{c}{Deblur-NeRF \cite{ma2022deblur}}  & \multicolumn{3}{c}{DP-NeRF \cite{dpnerf}} & \multicolumn{3}{c}{PDRF \cite{pdrf}} & \multicolumn{3}{c}{BAGS}  \\

     Motion& PSNR & SSIM & LPIPS & PSNR & SSIM & LPIPS & PSNR & SSIM & LPIPS & PSNR & SSIM & LPIPS & PSNR & SSIM & LPIPS & PSNR & SSIM & LPIPS \\

     \midrule[1pt]

    Ball & 24.08 & 0.624 & 0.399 & 23.22 & 0.619 & 0.340 & 27.36 & 0.766 & 0.223 & 27.20 & 0.765 & 0.209 & \tp{27.88} & 0.783 & 0.209 & 27.68 & \secp{0.799} & \fp{0.150}\\
    Basket & 23.72 & 0.709 & 0.322 & 23.24 & 0.688 & 0.288 & 27.67 & 0.845 & 0.148 & 27.74 & 0.846 & 0.129 & 28.63 & 0.869 & 0.119 & \tp{29.54} & \secp{0.900} & \fp{0.068} \\
    Buick & 21.59 & 0.633 & 0.350 & 21.46 & 0.658 & 0.266 & 24.77 & 0.770 & 0.175 & 25.70 & 0.792 & 0.141 & 25.69 & 0.790 & 0.165 & \tp{26.18} & \secp{0.844} & \fp{0.088} \\
    Coffee & 26.48 & 0.806 & 0.290 & 24.73 & 0.749 & 0.288 & 30.93 & 0.898 & 0.124 & \tp{32.44} & \secp{0.915} & 0.101 & 32.41 & 0.913 & 0.112 & 31.59 & 0.908 & \fp{0.096} \\
    Decoration & 22.39 & 0.661 & 0.363 & 20.55 & 0.641 & 0.299 & 24.19 & 0.771 & 0.186 & 23.51 & 0.740 & 0.275 & 23.29 & 0.734 & 0.228 & \tp{26.09} & \secp{0.858} & \fp{0.083} \\
    Girl & 20.07 & 0.708 & 0.320 & 19.87 & 0.714 & 0.278 & 22.27 & 0.798 & 0.169 & 21.80 & 0.770 & 0.205 & 23.94 & 0.830 & 0.171 & \tp{25.45} & \secp{0.869} & \fp{0.079} \\
    Heron & 20.50 & 0.522 & 0.413 & 19.43 & 0.505 & 0.332 & 22.63 & 0.687 & 0.210 & 22.52 & 0.676 & 0.281 & \tp{22.84} & 0.692 & 0.231 & 22.04 & \secp{0.715} & \fp{0.126} \\
    Parterre & 23.14 & 0.620 & 0.405 & 22.28 & 0.590 & 0.321 & 25.82 & 0.760 & 0.216 & 24.97 & 0.768 & 0.227 & 25.64 & 0.758 & 0.235 & \secp{25.92} & \secp{0.819} & \fp{0.092} \\
    Puppet & 22.09 & 0.609 & 0.339 & 22.05 & 0.631 & 0.267 & 25.24 & 0.751 & 0.158 & 25.12 & 0.758 & 0.181 & 24.99 & 0.761 & 0.147 & \tp{25.81} & \secp{0.804} & \fp{0.094} \\
    Stair & 22.87 & 0.456 & 0.487 & 21.91 & 0.474 & 0.387 & 25.39 & 0.630 & 0.210 & 25.78 & 0.647 & 0.229 & 25.73 & 0.639 & 0.211 & \tp{26.69} & \secp{0.721} & \fp{0.080} \\

    \midrule[1pt]
    
    Average & 22.69 & 0.635 & 0.369 & 21.87 & 0.627 & 0.306 & 25.63 & 0.767 & 0.182 & 25.91 & 0.775 & 0.160 & 26.10 & 0.777 & 0.183 & \tp{26.70} & \secp{0.824} & \fp{0.096} \\

     \midrule[1pt]

     Defocus  & & &  & & &  & & &  & & &  & & &  & & &  \\

     \midrule[1pt]
     
    Cake & 24.42 & 0.721 & 0.225 & 22.17 & 0.645 & 0.290 & 26.27 & 0.780 & 0.128 & 26.16 & 0.778 & 0.127 & 27.07 & 0.799 & 0.120 & \tp{27.21} & \secp{0.818} & \fp{0.108} \\
    Caps & 22.73 & 0.631 & 0.280 & 21.00 & 0.503 & 0.442 & 23.87 & 0.713 & 0.161 & 23.95 & 0.712 & \fp{0.143} & 24.10 & 0.716 & 0.147 & \tp{24.16} & \secp{0.725} & 0.159 \\
    Cisco & 20.72 & 0.722 & 0.126 & 20.08 & 0.709 & 0.163 & \tp{20.83} & 0.727 & 0.087 & 20.73 & 0.726 & 0.084 & 20.55 & 0.725 & 0.091 & 20.79 & \secp{0.743} & \fp{0.070} \\
    Coral & 19.81 & 0.566 & 0.269 & 19.60 & 0.555 & 0.323 & 19.85 & 0.600 & 0.121 & \tp{22.80} & \secp{0.741} & \fp{0.096} & 19.53 & 0.591 & 0.112 & 20.53 & 0.628 & 0.117 \\
    Cupcake & 21.88 & 0.681 & 0.216 & 21.55 & 0.681 & 0.213 & 22.26 & 0.722 & 0.116 & 20.11 & 0.611 & 0.118 & \tp{23.09} & 0.754 & 0.094 & 22.93 & \secp{0.762} & \fp{0.080} \\
    Cups & 25.02 & 0.758 & 0.232 & 20.93 & 0.646 & 0.314 & 26.21 & 0.799 & 0.127 & 26.75 & 0.814 & 0.104 & \tp{26.28} & 0.812 & 0.127 & 26.27 & \secp{0.823} & \fp{0.104} \\
    Daisy & 22.74 & 0.620 & 0.262 & 21.59 & 0.614 & 0.273 & 23.52 & 0.687 & 0.121 & 23.79 & 0.697 & 0.108 & \tp{24.39} & 0.741 & 0.091 & 23.74 & \secp{0.746} & \fp{0.062} \\
    
    Sausage & 17.79 & 0.483 & 0.279 & 17.78 & 0.473 & 0.284 & 18.01 & 0.500 & 0.180 & 18.35 & 0.544 & 0.147 & \tp{18.86} & 0.564 & 0.143 & 18.76 & \secp{0.574} & \fp{0.110} \\
    Seal & 22.79 & 0.627 & 0.268 & 22.13 & 0.592 & 0.310 & 26.04 & 0.777 & 0.105 & 25.95 & 0.778 & 0.103 & 26.47 & 0.806 & \fp{0.084} & \tp{26.52} & \secp{0.812} & 0.090 \\
    Tools & 26.08 & 0.852 & 0.155 & 23.98 & 0.812 & 0.196 & 27.81 & 0.895 & 0.061 & 28.07 & 0.898 & 0.054 & 28.14 & 0.901 & 0.059 & \tp{28.60} & \secp{0.913} & \fp{0.046} \\
    
    \midrule[1pt]
    Average & 22.40 & 0.666 & 0.231 & 21.08 & 0.623 & 0.281 & 23.47 & 0.720 & 0.121 & 23.67 & 0.730 & 0.108 & 23.85 & 0.741 & 0.107 & \tp{23.95} & \secp{0.754} & \fp{0.095} \\
    
    \end{tabular}
}    
    \caption{Quantitative comparisons on the camera motion and defocus blur dataset. We color code the best \tp{PSNR}, \secp{SSIM}, and \fp{LPIPS} performances.}
    \label{tab:results_blur}

\vspace{-2em}
\end{table*}

Finally, we ablate on the kernel dimension estimated by BPN. In Fig.~\ref{img:kernel_size}, we show that larger kernels in general lead to higher performances; interestingly, this is not true in the mixed resolution scenario. This is likely because downsacling is uniform across all pixels in an image, \ie a downscaling factor of four means a kernel size of five can fully explain the degradation. As such, increasing the kernel size does not improve performance and introduces noise instead.


\subsection{Quantitative Evaluation}
\myparagraph{Camera Motion and Defocus Blur}
As shown in Table~\ref{tab:results_blur}, we observe that Gaussian Splatting~\cite{mipsplatting} performs markedly worse than vanilla NeRF in the presence of blurry images, despite its strong performance in other curated benchmarks. Since NeRF models a continuous neural volume with an MLP, individual degradation can be smoothly explained by spreading the error into empty space. On the other hand, 3DGS is based on a discrete representation and tends to produce concentrated artifacts. Our BAGS can achieve new SoTA performances compared to previous NeRF-based deblurring methods despite the unfavorable performances in 3DGS. Specifically, we find significant improvements on camera motion blur across all metrics and achieve a \ul{fifty percent improvement in LPIPS} on average. As visualized in Fig.~\ref{tab:visualization}, BAGS achieves much sharper results compared to previous methods~\cite{ma2022deblur,dpnerf,exblurf,pdrf}, to the point where it is hard to distinguish the groundtruth image from our rendered results. 

While improvements on defocus blur is less drastic in Table~\ref{tab:results_blur}, we still observe great visual improvements. Several reasons contribute to the less apparent quantitative improvements. First, the collected groundtruth images have different exposure compared to training images; as such, image-wise similarity metrics like PSNR are less reliable. Secondly, many scenes contain specular surfaces, which all methods do a bad job at reconstructing. As shown in Fig.~\ref{tab:visualization}, much better details can be recovered through BAGS than other methods.




\captionsetup[subfigure]{labelformat=empty}
\begin{figure*}[htb!]
    \setlength{\tabcolsep}{0.5pt}
    \centering
    \tiny

    \begin{tabular}[b]{ccccccccc}
    
         \tiny{Novel View} & \tiny{NeRF} & \tiny{Mip-Sp} & \tiny{Db-NeRF} & \tiny{DP-NeRF} & \tiny{ExBluRF} & \tiny{PDRF} & \tiny{Ours} & \tiny{G.T.}\\


        \begin{subfigure}[b]{0.19\linewidth}
            \includegraphics[width=\textwidth]{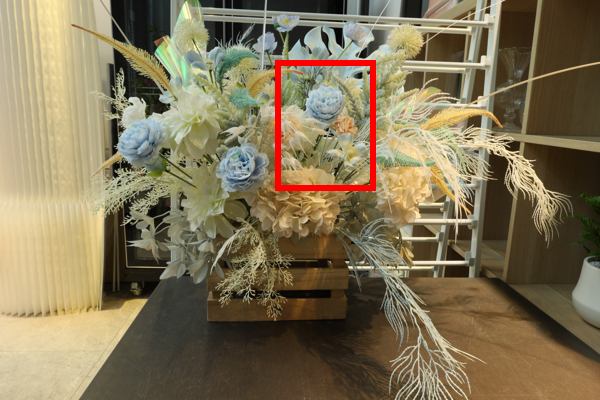}
            \caption{\tiny{PSNR/SSIM}}
        \end{subfigure} 
        &  
        \begin{subfigure}[b]{0.097\linewidth}
            \includegraphics[width=\textwidth]{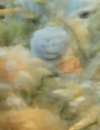}
            \caption{\tiny{22.6/.621}}
        \end{subfigure} &
        \begin{subfigure}[b]{0.097\linewidth}
            \includegraphics[width=\textwidth]{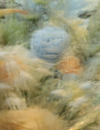}
            \caption{\tiny{22.5/.658}}
        \end{subfigure} 
        &
        \begin{subfigure}[b]{0.097\linewidth}
            \includegraphics[width=\textwidth]{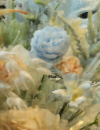}
            \caption{\tiny{25.35/.815}}
        \end{subfigure} 
        &
        \begin{subfigure}[b]{0.097\linewidth} 
            \includegraphics[width=\textwidth]{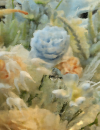}
            \caption{\tiny{{25.54/.826}}}
        \end{subfigure} &
        \begin{subfigure}[b]{0.097\linewidth}
            \includegraphics[width=\textwidth]{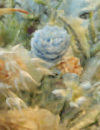}
            \caption{\tiny{23.39/.761}}
        \end{subfigure} &  
        \begin{subfigure}[b]{0.097\linewidth}
            \includegraphics[width=\textwidth]{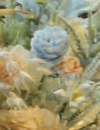}
            \caption{\tiny{24.87/.793}}
        \end{subfigure} &
        \begin{subfigure}[b]{0.097\linewidth}
            \includegraphics[width=\textwidth]{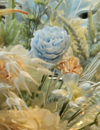}
            \caption{\tiny{\tp{28.74}/\secp{.934}}}
        \end{subfigure} &
        \begin{subfigure}[b]{0.097\linewidth}
            \includegraphics[width=\textwidth]{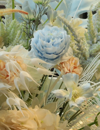}
            \caption{\tiny{Motion}}
        \end{subfigure}\\

        \begin{subfigure}[b]{0.19\linewidth}
            \includegraphics[width=\textwidth]{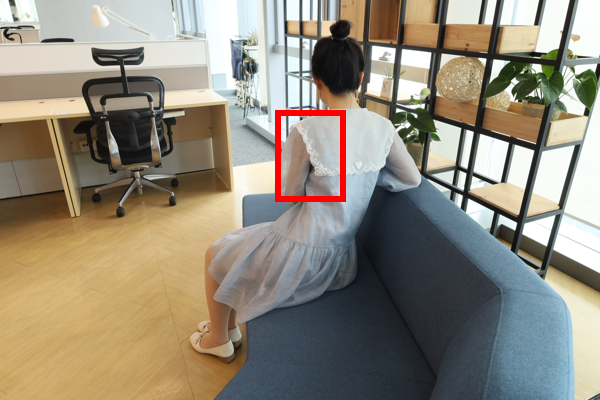}
            \caption{\tiny{PSNR/SSIM}}
        \end{subfigure} 
        &  
        \begin{subfigure}[b]{0.097\linewidth}
            \includegraphics[width=\textwidth]{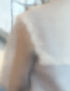}
            \caption{\tiny{23.29/.814}}
        \end{subfigure}&
        \begin{subfigure}[b]{0.097\linewidth}
            \includegraphics[width=\textwidth]{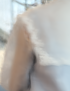}
            \caption{\tiny{23.27/.845}}
        \end{subfigure} 
        &
        \begin{subfigure}[b]{0.097\linewidth}
            \includegraphics[width=\textwidth]{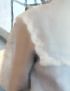}
            \caption{\tiny{27.08/.902}}
        \end{subfigure} 
        &
        \begin{subfigure}[b]{0.097\linewidth}
            \includegraphics[width=\textwidth]{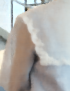}
            \caption{\tiny{{27.54/.908}}}
        \end{subfigure}&
        \begin{subfigure}[b]{0.097\linewidth}
            \includegraphics[width=\textwidth]{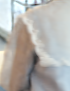}
            \caption{\tiny{25.66/.892}}
        \end{subfigure}&  
        \begin{subfigure}[b]{0.097\linewidth}
            \includegraphics[width=\textwidth]{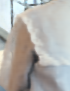}
            \caption{\tiny{{26.66/.898}}}
        \end{subfigure} &
        \begin{subfigure}[b]{0.097\linewidth}
            \includegraphics[width=\textwidth]{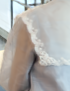}
            \caption{\tiny{\tp{28.03}/\secp{.927}}}
        \end{subfigure} &
        \begin{subfigure}[b]{0.097\linewidth}
            \includegraphics[width=\textwidth]{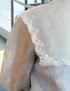}
            \caption{\tiny{Motion}}
        \end{subfigure}\\ 
        
    \end{tabular}

    \vspace{3pt}

    \begin{tabular}[b]{cccccccc}
         \tiny{Novel View} & \tiny{NeRF} & \tiny{Mip-Sp} & \tiny{Db-NeRF} & \tiny{DP-NeRF} & \tiny{PDRF} & \tiny{Ours} & \tiny{G.T.}\\
        
        \begin{subfigure}[b]{0.21\linewidth}
            \includegraphics[width=\textwidth]{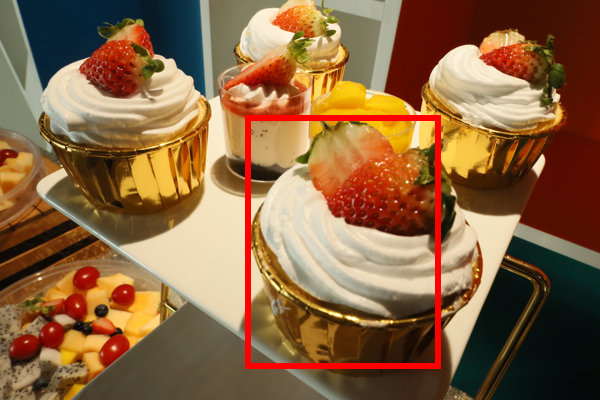}
            \caption{\tiny{PSNR/SSIM}}
        \end{subfigure} 
        &  
        \begin{subfigure}[b]{0.108\linewidth}
            \includegraphics[width=\textwidth]{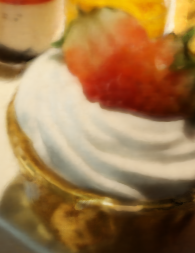}
            \caption{\tiny{22.16/.709}}
        \end{subfigure}
        &
        \begin{subfigure}[b]{0.108\linewidth}
            \includegraphics[width=\textwidth]{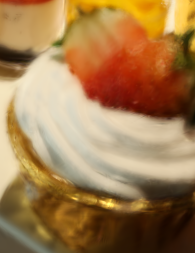}
            \caption{\tiny{22.23/.709}}
        \end{subfigure} 
        &
        \begin{subfigure}[b]{0.108\linewidth}
            \includegraphics[width=\textwidth]{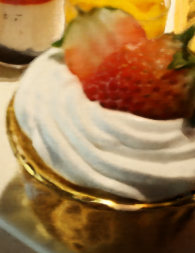}
            \caption{\tiny{23.14/.765}}
        \end{subfigure} 
        &
        \begin{subfigure}[b]{0.108\linewidth}
            \includegraphics[width=\textwidth]{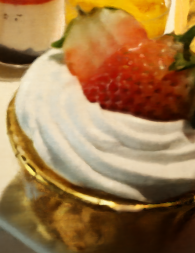}
            \caption{\tiny{25.24/.822}}
        \end{subfigure}
        &   
        \begin{subfigure}[b]{0.108\linewidth}
            \includegraphics[width=\textwidth]{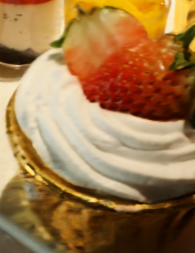}
            \caption{\tiny{{25.47/.833}}}
        \end{subfigure} 
        &
        \begin{subfigure}[b]{0.108\linewidth}
            \includegraphics[width=\textwidth]{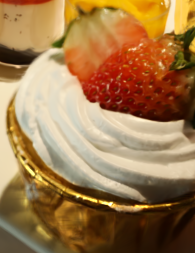}
            \caption{\tiny{\tp{26.14}/\secp{.879}}}
        \end{subfigure} 
        &
        \begin{subfigure}[b]{0.108\linewidth}
            \includegraphics[width=\textwidth]{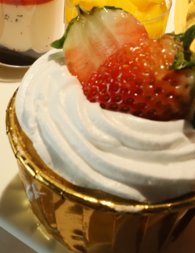}
            \caption{\tiny{Defocus}}
        \end{subfigure}\\ 

        \begin{minipage}[b]{0.21\linewidth}
            \includegraphics[width=\textwidth]{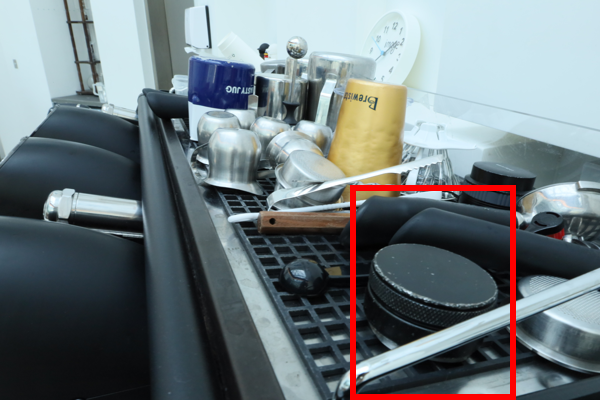}
            \captionsetup{labelformat=empty,skip=0pt}
            \caption{\tiny{PSNR/SSIM}}
        \end{minipage} 
        &  
        \begin{minipage}[b]{0.108\linewidth}
            \includegraphics[width=\textwidth]{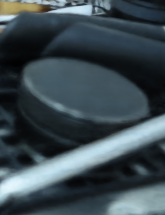}
            \captionsetup{labelformat=empty,skip=0pt}
            \caption{\tiny{22.91/.774}}
        \end{minipage}
        &
        \begin{minipage}[b]{0.108\linewidth}
            \includegraphics[width=\textwidth]{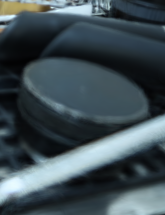}
            \captionsetup{labelformat=empty,skip=0pt}
            \caption{\tiny{21.74/.758}}

        \end{minipage} 
        &
        \begin{minipage}[b]{0.108\linewidth}
            \includegraphics[width=\textwidth]{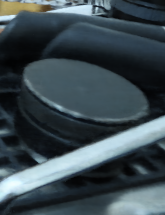}
            \captionsetup{labelformat=empty,skip=0pt}
            \caption{\tiny{24.98/.856}}
        \end{minipage} 
        &
        \begin{minipage}[b]{0.108\linewidth}
            \includegraphics[width=\textwidth]{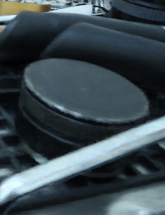}
            \captionsetup{labelformat=empty,skip=0pt}
            \caption{\tiny{25.38/.857}}
        \end{minipage}
        &   
        \begin{minipage}[b]{0.108\linewidth}
            \includegraphics[width=\textwidth]{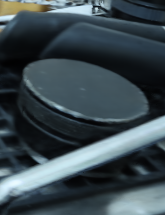}
            \captionsetup{labelformat=empty,skip=0pt}
            \caption{\tiny{{26.15/.881}}}
        \end{minipage} 
        &
        \begin{minipage}[b]{0.108\linewidth}
            \includegraphics[width=\textwidth]{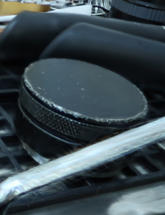}
            \captionsetup{labelformat=empty,skip=0pt}
            \caption{\tiny{\tp{26.97}/\secp{.907}}}
        \end{minipage} 
        &
        \begin{minipage}[b]{0.108\linewidth}
            \includegraphics[width=\textwidth]{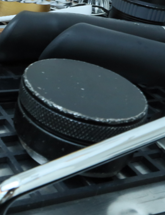}
            \captionsetup{labelformat=empty,skip=0pt}
            \caption{\tiny{Defocus}}
        \end{minipage} \\ 

    \end{tabular}

    \setcounter{figure}{4}
    \caption{Visualizations of test views on camera motion and defocus blur dataset. Mip-Sp and Db-NeRF are short for Mip-Splatting \cite{mipsplatting} and Deblur-NeRF \cite{ma2022deblur}.}
    \label{tab:visualization}
    \vspace{-2em}

\end{figure*}

\myparagraph{Mix Resolution and Low Light Motion Blur}
We investigate the robustness of several SoTA reconstruction methods on downscaling perturbation and motion blur observed in low light condition. 
The experiments are done on \ul{unbounded scenes} and are more challenging and realistic than the forward bounded scenes in DeblurNeRF~\cite{ma2022deblur}. We note that previous approaches~\cite{ma2022deblur,dpnerf,badnerf,pdrf} cannot handle unbounded scenes; therefore, we do not include them in Table~\ref{tab:mixres_lowlight}.


As shown in Table~\ref{tab:mixres_lowlight}, Mip Splatting performs on par with Mip-NeRF 360; both are vastly better than NeRFacto~\cite{nerfstudio}, which stores hash features~\cite{mueller2022instant} to accelerate the training process and uses a very small MLP represent the scene. We also employ a Super-Resolution (SR) model HAT~\cite{hat} on downscaled images in the hope that single-image SR can improve performances. However, pre-trained SR models are limited by domain shifts, dedicated upscaling factors, and the lack of multi-view consistency. As visualized in Fig.~\ref{tab:visualization_sr}, while the render is somewhat sharper after SR, the color becomes unfaithful. Compared with all other methods, our rendered results are much sharper in details. This robust performance comes from BPN's ability to model Gaussian blur on high resolution before comparing them with the downscaled observations.

Motion blur is often observed in low light condition. To this end, we perform experiments on scenes acquired near sunset or at night. As shown in Table~\ref{tab:mixres_lowlight}, BAGS again improves significantly over other reconstruction approaches, which shows that our method can be generalized to different imaging geometry. As shown in Fig.~\ref{tab:visualization_sr}, we employ MPR~\cite{Zamir2021MPRNet}, a single image deblurring model, on the motion-blurred images before reconstructing with Mip-Splatting~\cite{mipsplatting}. Similar to previous attempts, the deblurring prior leads to a slightly clearer rendering but is not sufficient to remove all blur. BAGS reconstructs the scene remarkably well compared to other methods, where even the window details can be recovered correctly. Please refer to the supplemental material for more visualizations.

\begin{table*}[htbp]
    \centering
    \small
    \setlength{\tabcolsep}{0.5pt}
    \resizebox{\linewidth}{!}{

    \begin{tabular}{l | ccccccccccccccc}

     \multicolumn{1}{c|}{\multirow{2}[0]{*}{MixedRes}}  & \multicolumn{3}{c}{NeRFacto \cite{nerfstudio}}  & \multicolumn{3}{c}{Mip-NeRF 360 \cite{barron2022mipnerf360}} &  \multicolumn{3}{c}{Mip-Sp \cite{mipsplatting}}  & \multicolumn{3}{c}{Mip-Sp-HAT~\cite{hat}} & \multicolumn{3}{c}{BAGS}  \\

     & PSNR & SSIM & LPIPS & PSNR & SSIM & LPIPS & PSNR & SSIM & LPIPS & PSNR & SSIM & LPIPS & PSNR & SSIM & LPIPS \\

     \midrule[1pt]

    Bicycle & 19.25 & 0.349 & 0.496 & 23.82 & 0.610 & 0.375 & 24.03 & 0.628 & 0.364 & 23.20 & 0.559 & 0.408 & \tp{24.97} & \secp{0.724} & \fp{0.251} \\
    Bonsai & 21.80 & 0.619 & 0.259 & 29.91 & 0.871 & 0.190 & 29.27 & 0.876 & 0.193 & 28.38 & 0.848 & 0.228 & \tp{31.11} & \secp{0.933} & \fp{0.116} \\
    Counter & 24.07 & 0.714 & 0.280 & 28.09 & 0.848 & 0.220 & 27.79 & 0.853 & 0.216 & 27.02 & 0.811 & 0.252 & \tp{29.03} & \secp{0.900} & \fp{0.140} \\
    Flowers & 19.63 & 0.418 & 0.459 & 21.98 & 0.566 & 0.389 & 21.96 & 0.570 & 0.397 & 20.81 & 0.520 & 0.415 & \tp{22.47} & \secp{0.633} & \fp{0.330} \\
    Garden & 21.61 & 0.489 & 0.357 & 25.34 & 0.667 & 0.317 & 25.28 & 0.682 & 0.305 & 23.96 & 0.611 & 0.364 & \tp{26.45} & \secp{0.800} & \fp{0.178} \\
    Kitchen & 22.73 & 0.584 & 0.304 & 27.82 & 0.779 & 0.238 & 27.49 & 0.778 & 0.241 & 26.10 & 0.730 & 0.296 & \tp{30.86} & \secp{0.925} & \fp{0.095} \\
    Room & 24.68 & 0.791 & 0.214 & 30.38 & 0.902 & 0.198 & 30.43 & 0.903 & 0.203 & 29.61 & 0.240 & \fp{0.135} & \tp{30.50} & \secp{0.907} & 0.169 \\
    Stump & 23.19 & 0.555 & 0.377 & 25.75 & 0.688 & 0.320 & 25.87 & 0.703 & 0.299 & 24.09 & 0.618 & 0.364 & \tp{25.97} & \secp{0.720} & \fp{0.258} \\
    Treehill & 20.11 & 0.416 & 0.501 & 22.52 & 0.558 & 0.437 & \tp{22.86} & 0.550 & 0.457 & 21.76 & 0.494 & 0.482 & 22.60 & \secp{0.579} & \fp{0.389} \\

    \midrule[1pt]
    Average & 21.90 & 0.548 & 0.361 & 26.18 & 0.721 & 0.298 & 26.11 & 0.727 & 0.297 & 24.99 & 0.603 & 0.327 & \tp{27.11} & \secp{0.791} & \fp{0.214} \\



    \midrule[1pt]

    \multicolumn{1}{c|}{\multirow{2}[0]{*}{Lowlight-drone}}  & \multicolumn{3}{c}{NeRFacto \cite{nerfstudio}} & \multicolumn{3}{c}{Mip-NeRF 360 \cite{barron2022mipnerf360}} &  \multicolumn{3}{c}{Mip-Sp \cite{mipsplatting}}  & \multicolumn{3}{c}{Mip-Sp-MPR~\cite{Zamir2021MPRNet}} & \multicolumn{3}{c}{BAGS} \\
     & PSNR & SSIM & LPIPS & PSNR & SSIM & LPIPS & PSNR & SSIM & LPIPS & PSNR & SSIM & LPIPS & PSNR & SSIM & LPIPS \\

    \midrule[1pt]

    Math-night & 21.33 & 0.582 & 0.326 & 29.51 & 0.882 & 0.191 & 29.61 & 0.884 & 0.199 & 29.15 & 0.877 & 0.204 & \tp{30.75}&\secp{0.905}&\fp{0.158}\\
    Library-night  & 24.48 & 0.768 & 0.381 & 26.38 & 0.821 & 0.349 & 26.17 & 0.820 & 0.370 & 25.89 & 0.810 & 0.385 & \tp{28.55}&\secp{0.859}&\fp{0.313}\\
    BioMed-sunset & 23.76 & 0.650 & 0.439 & 24.48 & 0.688 & 0.402 & 24.85 & 0.712 & 0.426 & 24.78 & 0.702 & 0.427 & \tp{28.67}&\secp{0.854}&\fp{0.201}\\
    Admin-sunset & 21.96 & 0.621 & 0.400 & 25.47 & 0.757 & 0.355 & 25.54 & 0.764 & 0.369 & 25.19 & 0.751 & 0.373 & \tp{29.01}&\secp{0.855}&\fp{0.216}\\
    Aud-sunset & 24.67 & 0.709 & 0.398 & 25.90 & 0.750 & 0.368 & 26.33 & 0.770 & 0.386 & 26.77 & 0.785 & 0.359 & \tp{32.34}&\secp{0.914}&\fp{0.155}\\

    \midrule[1pt]
    Average & 23.24 & 0.666 & 0.389 & 26.35 & 0.780 & 0.333 & 26.50 & 0.790 & 0.350 & 26.36 & 0.785 & 0.350 & \tp{29.86} & \secp{0.877} & \fp{0.209} \\

\end{tabular}
}
\caption{Quantitative comparisons on mixed resolution and low light motion blur in unbounded 360 geometry. Mip-Sp is short for Mip-Splatting \cite{mipsplatting}. Mip-Sp-HAT and Mip-Sp-HAT-MPR use prior models to restore images first before Mip-Splatting.}
\label{tab:mixres_lowlight}
\vspace{-2em}
\end{table*}






\begin{figure*}[htb!]

    \setlength{\abovecaptionskip}{3pt}
    \setlength{\tabcolsep}{0.5pt}
    \centering
    
    \begin{tabular}[b]{ccccccc}
    
         \tiny{Novel View} & \tiny{NeRFacto} & \tiny{MipNeRF-360} & \tiny{Mip-Sp} & \tiny{Mip-Sp-HAT} & \tiny{Ours} & \tiny{G.T.}\\

        \begin{subfigure}[b]{0.128\linewidth}
            \includegraphics[width=\textwidth]{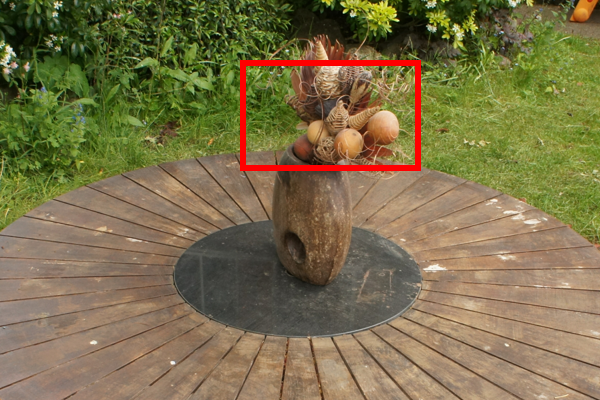}
            \caption{\tiny{PSNR/SSIM}}
        \end{subfigure} 
        &  
        \begin{subfigure}[b]{0.14\linewidth}
            \includegraphics[width=\textwidth]{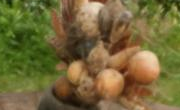}
            \caption{\tiny{21.61/.576}}
        \end{subfigure} &
        \begin{subfigure}[b]{0.14\linewidth}
            \includegraphics[width=\textwidth]{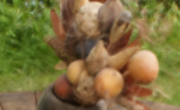}
            \caption{\tiny{{24.09/.670}}}
        \end{subfigure} 
        &
        \begin{subfigure}[b]{0.14\linewidth}
            \includegraphics[width=\textwidth]{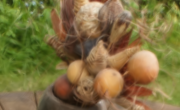}
            \caption{\tiny{{25.11/.746}}}
        \end{subfigure} &  
        \begin{subfigure}[b]{0.14\linewidth}
            \includegraphics[width=\textwidth]{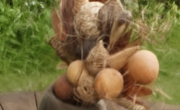}
            \caption{\tiny{24.41/.722}}
        \end{subfigure} &
        \begin{subfigure}[b]{0.14\linewidth}
            \includegraphics[width=\textwidth]{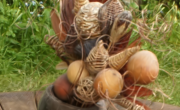}
            \caption{\tiny{\tp{28.46}/\secp{.899}}}
        \end{subfigure} &
        \begin{subfigure}[b]{0.14\linewidth}
            \includegraphics[width=\textwidth]{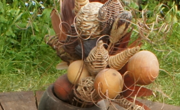}
            \caption{\tiny{MixedRes}}
        \end{subfigure}\\

        \begin{subfigure}[b]{0.13\linewidth}
            \includegraphics[width=\textwidth]{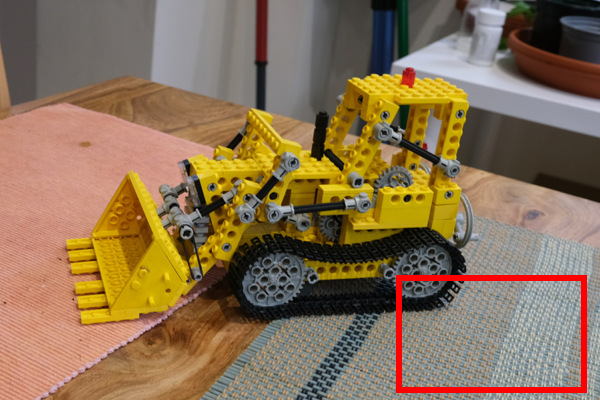}
            \caption{\tiny{PSNR/SSIM}}
        \end{subfigure} 
        &  
        \begin{subfigure}[b]{0.14\linewidth}
            \includegraphics[width=\textwidth]{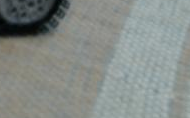}
            \caption{\tiny{23.08/.412}}
        \end{subfigure}&
        \begin{subfigure}[b]{0.14\linewidth}
            \includegraphics[width=\textwidth]{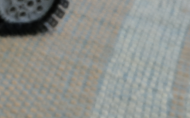}
            \caption{\tiny{{24.98/.562}}}
        \end{subfigure} 
        &
        \begin{subfigure}[b]{0.14\linewidth}
            \includegraphics[width=\textwidth]{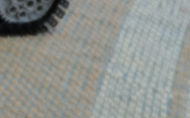}
            \caption{\tiny{{24.93/.590}}}
        \end{subfigure} 
        &
        \begin{subfigure}[b]{0.14\linewidth}
            \includegraphics[width=\textwidth]{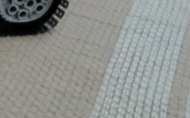}
            \caption{\tiny{24.68/.587}}
        \end{subfigure} &
        \begin{subfigure}[b]{0.14\linewidth}
            \includegraphics[width=\textwidth]{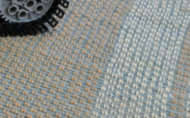}
            \caption{\tiny{\tp{31.79}/\secp{.949}}}
        \end{subfigure} &
        \begin{subfigure}[b]{0.14\linewidth}
            \includegraphics[width=\textwidth]{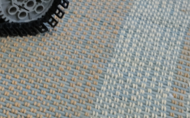}
            \caption{\tiny{MixedRes}}
        \end{subfigure}\\ 
        
    \end{tabular}

    \begin{tabular}[b]{ccccccc}
    
         \tiny{Novel View} & \tiny{NeRFacto} & \tiny{MipNeRF-360} & \tiny{Mip-Sp} & \tiny{Mip-Sp-MPR} & \tiny{Ours} & \tiny{G.T.}\\

        \begin{subfigure}[b]{0.128\linewidth}
            \includegraphics[width=\textwidth]{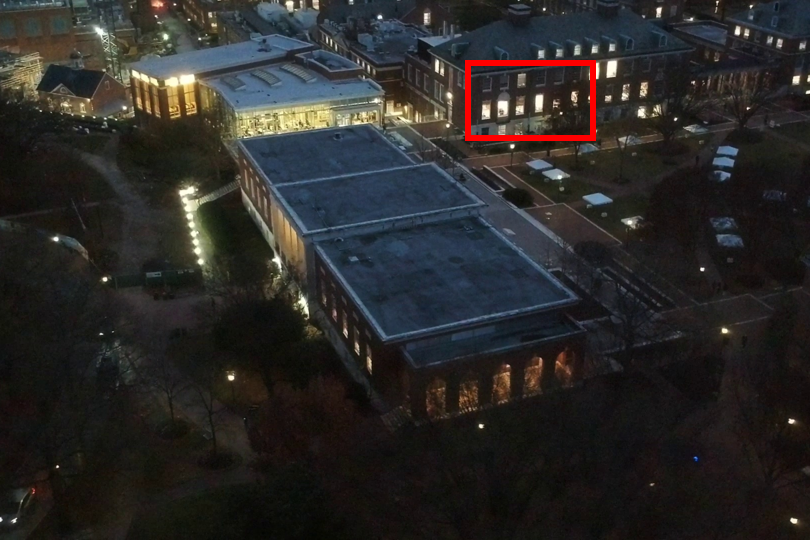}
            \caption{\tiny{PSNR/SSIM}}
        \end{subfigure} 
        &  
        \begin{subfigure}[b]{0.14\linewidth}
            \includegraphics[width=\textwidth]{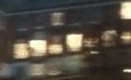}
            \caption{\tiny{17.86/.608}}
        \end{subfigure} &
        \begin{subfigure}[b]{0.14\linewidth}
            \includegraphics[width=\textwidth]{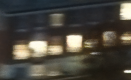}
            \caption{\tiny{18.33/.704}}
        \end{subfigure} 
        &
        \begin{subfigure}[b]{0.14\linewidth}
            \includegraphics[width=\textwidth]{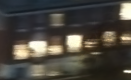}
            \caption{\tiny{{18.57/.714}}}
        \end{subfigure} &  
        \begin{subfigure}[b]{0.14\linewidth}
            \includegraphics[width=\textwidth]{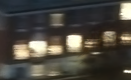}
            \caption{\tiny{{18.51/.718}}}
        \end{subfigure} &
        \begin{subfigure}[b]{0.14\linewidth}
            \includegraphics[width=\textwidth]{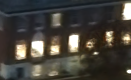}
            \caption{\tiny{\tp{24.47}/\secp{.850}}}
        \end{subfigure} &
        \begin{subfigure}[b]{0.14\linewidth}
            \includegraphics[width=\textwidth]{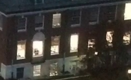}
            \caption{\tiny{LowLight}}
        \end{subfigure}\\ 
        
        \begin{subfigure}[b]{0.13\linewidth}
            \includegraphics[width=\textwidth]{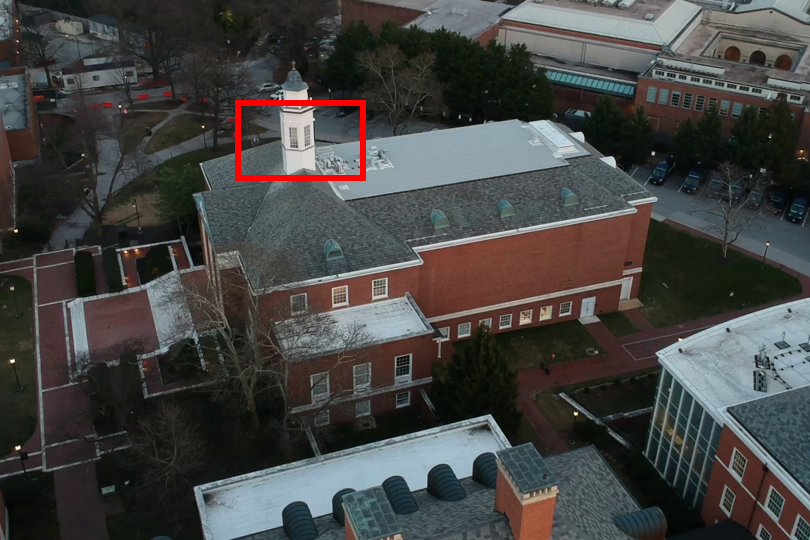}
            \caption{\tiny{PSNR/SSIM}}
        \end{subfigure} 
        &  
        \begin{subfigure}[b]{0.14\linewidth}
            \includegraphics[width=\textwidth]{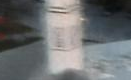}
            \caption{\tiny{22.40/.683}}
        \end{subfigure}&
        \begin{subfigure}[b]{0.14\linewidth}
            \includegraphics[width=\textwidth]{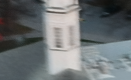}
            \caption{\tiny{25.36/.813}}
        \end{subfigure} 
        &
        \begin{subfigure}[b]{0.14\linewidth}
            \includegraphics[width=\textwidth]{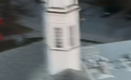}
            \caption{\tiny{25.51/.817}}
        \end{subfigure} 
        &
        \begin{subfigure}[b]{0.14\linewidth}
            \includegraphics[width=\textwidth]{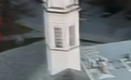}
            \caption{\tiny{{27.03/.867}}}
        \end{subfigure} &
        \begin{minipage}[b]{0.14\linewidth}
            \includegraphics[width=\textwidth]{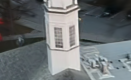}
            \captionsetup{labelformat=empty,skip=1pt}
            \caption{\tiny{\tp{31.75}/\secp{.938}}}
        \end{minipage} &
        \begin{subfigure}[b]{0.14\linewidth}
            \includegraphics[width=\textwidth]{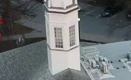}
            \caption{\tiny{LowLight}}
        \end{subfigure}\\ 
        
    \end{tabular}

    \setcounter{figure}{6}
    \caption{Visualizations of test views on mixed resolution and low light motion blur dataset. }
    \label{tab:visualization_sr}

\end{figure*}

\begin{figure}[!htb]
    \setlength{\abovecaptionskip}{3pt}
    \setlength{\tabcolsep}{1pt}
    \begin{tabular}[b]{cccc}
        \begin{subfigure}[b]{.245\linewidth}
            \includegraphics[width=\textwidth]{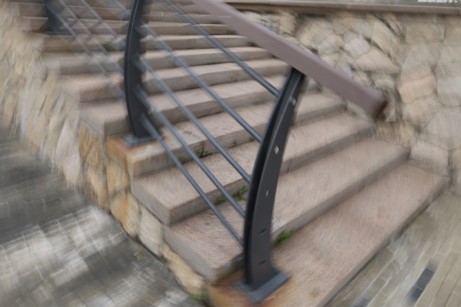}
            \caption{Camera Motion}
            \label{motion}
        \end{subfigure} &
        \begin{subfigure}[b]{.245\linewidth}
            \includegraphics[width=\textwidth]{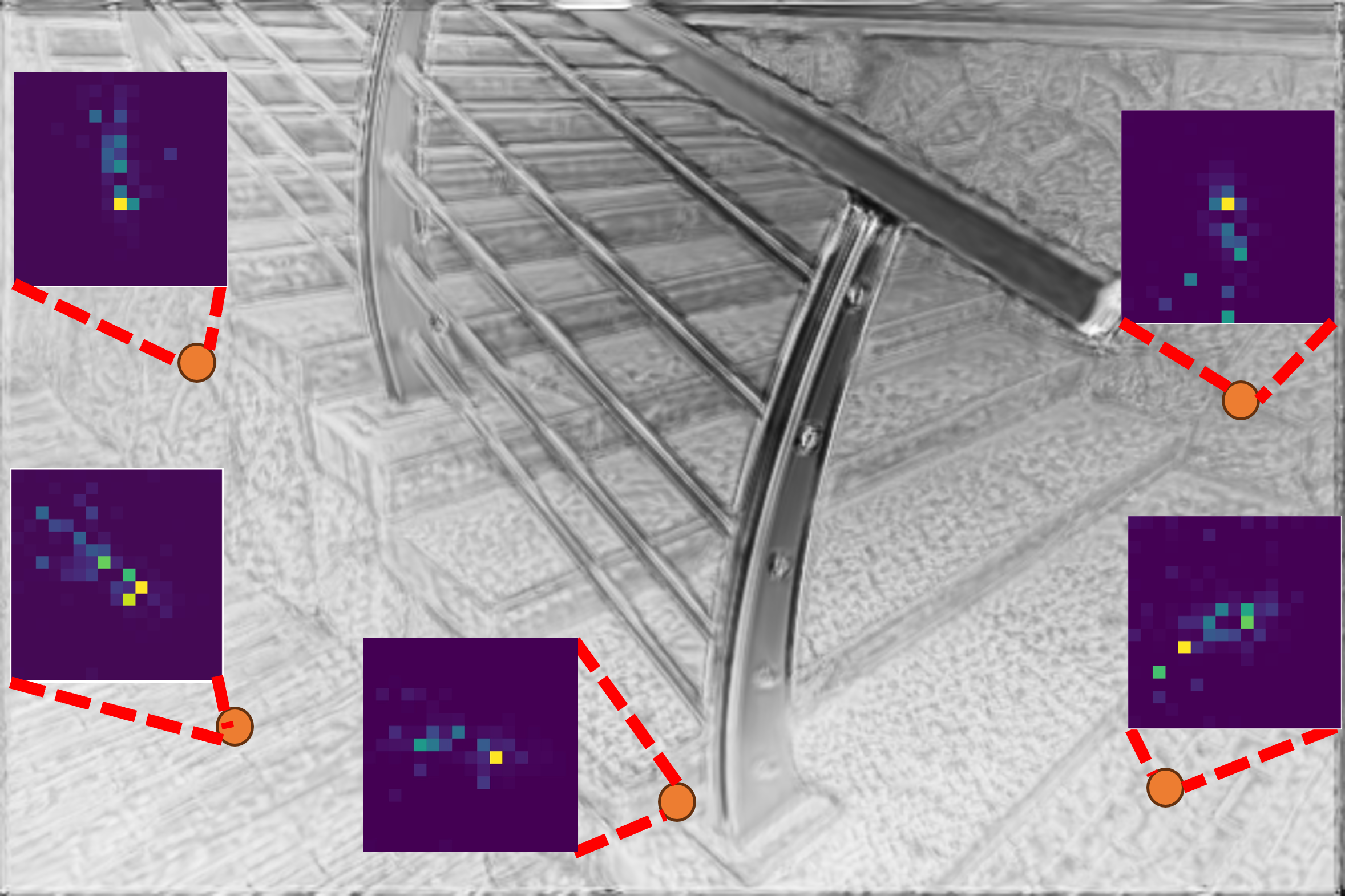}
            \caption{Kernel \& mask}
            \label{motion_kernel}
        \end{subfigure} &
        \begin{subfigure}[b]{.245\linewidth}
            \includegraphics[width=\textwidth]{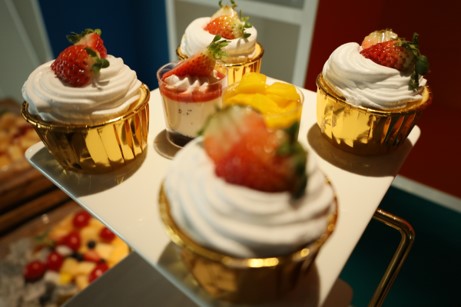}
            \caption{Defocus}
            \label{defocus}
        \end{subfigure} &
        \begin{subfigure}[b]{.245\linewidth}
            \includegraphics[width=\textwidth]{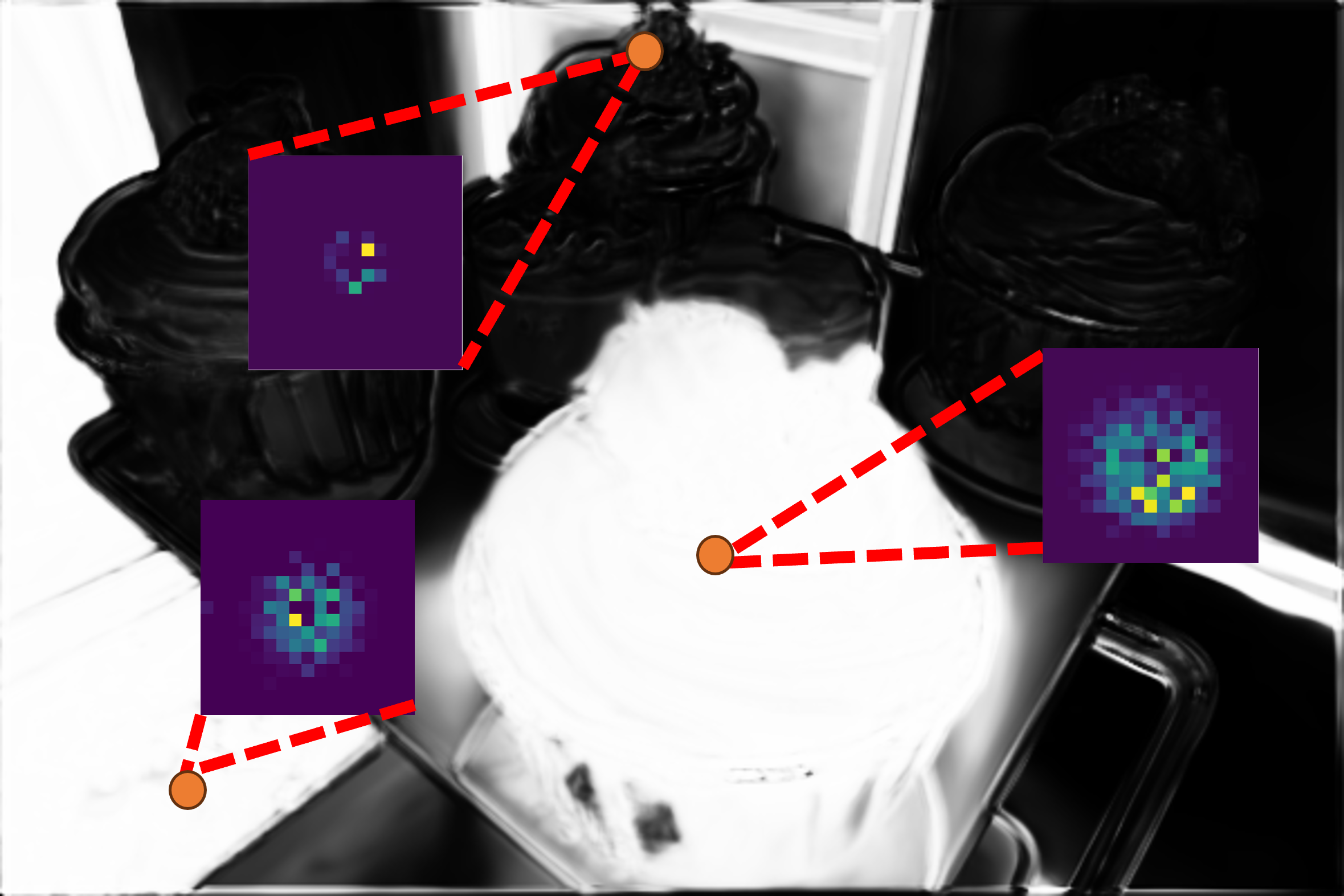}
            \caption{Kernel \& mask}
            \label{defocus+vis}
        \end{subfigure}
    \end{tabular}
    \caption{Visualization of estimated convolution kernels and masks in different blur.}
    \label{fig:kernel_mask_vis}
    \vspace{-1.5em}
\end{figure}

\myparagraph{Kernel and Mask Visualization} BAGS models image degradation through convolution kernels $h$ and masks $m$, which we visualize in Fig.~\ref{fig:kernel_mask_vis}. By constraining on its sparsity, we can see that $m$ meaningfully highlights the regions of blur in a given training image. On the other hand, low-value regions indicate either high quality observation or textureless regions. By visualizing the kernels modeled at different pixel, we can also easily characterize the types of observed blur. Specifically, the estimated kernels in camera motion blur exhibit clear patterns of camera movement, while those for defocus blur show Gaussian-like distributions based on the pixel's distance from the focus plane. These self-emerged properties in BAGS provide ways for us to automatically and precisely evaluate the quality of training images.

\vspace{-0.5em}

\myparagraph{Discussion}
 BAGS can achieve great results; however, there also exists many potential improvements. The added neural networks and convolution operations require additional computation. While BAGS is a generalist approach and addresses multiple types of blur, it may become very expensive to estimate per-pixel convolution in high resolution. To this end, we have explored utilizing pixel shuffling~\cite{shi2016real}, which converts spatial resolution into channels, before we estimate blur kernel; \ie, this assumes that within a small region, blur kernel $h$ stays roughly consistent. As shown in the supplemental material, we can scale to great visual results at 2K resolution on the DeblurNeRF dataset. Many other potential directions can also address this complexity issue, including leveraging the mask $m$ to identify important regions to focus on, a better designed low rank kernel estimation, degradation-specific optimization, \etc. As such, while BAGS is already more efficient than comparable NeRF-based methods, we anticipate that its computation cost can be further optimized in the future.

\section{Conclusion}
We present BAGS, a novel scene reconstruction method that can handle various noise in input images. We demonstrate that a vanilla Gaussian Splatting is particularly susceptible to degraded images due to its discrete representation, even compared with NeRF. To address this issue, BAGS implements a 2D degradation model, BPN, which estimates convolution kernels jointly with 3D scene optimization. BPN allows for additional freedom to address 3D inconsistencies, and produces interpretable kernels and masks to indicate the degradation types and regions within an image. Additionally, BAGS leverages a coarse-to-fine kernel optimization scheme, which gradually models the 3D scene with higher resolution images and kernels. This approach successfully addresses the 2D-3D ambiguity in optimizing convolution kernels with a very sparse point cloud. We perform extensive experiments on a variety of noise in training images, including camera motion blur, defocus blur, and low resolution. Under these conditions, BAGS can render high fidelity scenes and significantly improves upon previous approaches. In the future, we hope to further improve the computational efficiency in BAGS and dynamically adjust kernel capacity based on how degraded the training views are.


%
%
\bibliographystyle{splncs04}
\bibliography{main}
\end{document}